\documentclass[final,5p,times,twocolumn]{elsarticle}

\usepackage{colortbl}

\newcommand{\etal}{\emph{et al.~}} 
\newcommand{\tbf}{\textbf}

\newcommand{\mbr}[1]{\mathrm{\mathbf{#1}}}
\newcommand{\mbg}[1]{\boldsymbol{#1}}
\newcommand{\mr}[1]{\mathrm{#1}}
\newcommand{\para}[1]{\paragraph{#1}}
\newcommand{\itemtitle}[1]{\emph{#1}}
\newcommand{\diag}{\text{\textsf{diag}}}
\newcommand{\wrt}{w.r.t.~}
\newcommand{\nvs}{\vspace{-0.2cm}}
\newcommand{\indicator}{\mathbbm{1}}
\usepackage{amssymb,amsmath,algorithm2e,bbm,slashbox}
\journal{Computer Vision and Image Understanding}

\begin{document}
	
	\begin{frontmatter}
		
		
		
		\title{ASIST: Automatic Semantically Invariant Scene Transformation}
		
		
		\author[MicrosoftA,TelAviv]{Or Litany}
		\ead{t-orlita@microsoft.com}
		
		\author[MicrosoftA,TelAviv]{Tal Remez}
		\ead{t-talrem@microsoft.com}
		
		\author[MicrosoftA]{Daniel Freedman}
		\ead{danifree@microsoft.com}
		
		\author[MicrosoftB]{Lior Shapira}
		\ead{liors@microsoft.com}
		
		\author[TelAviv]{Alex Bronstein}
		\ead{bron@eng.tau.ac.il}
		
		\author[MicrosoftB]{Ran Gal}
		\ead{rgal@microsoft.com}
		
		\address[MicrosoftA]{Microsoft Research, Haifa, Israel }
		
		\address[TelAviv]{Electrical Engineering Department, Tel Aviv University, Tel Aviv, Israel}
		
		\address[MicrosoftB]{Microsoft Research, Redmond, WA, USA}
		
		\begin{abstract}
			
			We present ASIST, a technique for transforming point clouds by replacing objects with their semantically equivalent counterparts.  Transformations of this kind have applications in virtual reality, repair of fused scans, and robotics.  ASIST is based on a unified formulation of semantic labeling and object replacement; both result from minimizing a single objective.  We present numerical tools for the efficient solution of this optimization problem.  The method is experimentally assessed on new datasets of both synthetic and real point clouds, and is additionally compared to two recent works on object replacement on data from the corresponding papers.
			
		\end{abstract}
		
		
	\end{frontmatter}
	
	
	\section{Introduction}
	\label{sec:Introduction}
\begin{figure}
		\label{teaser}
		\centering
		\begin{tabular}{c c}
			\includegraphics[width = 0.23\textwidth]{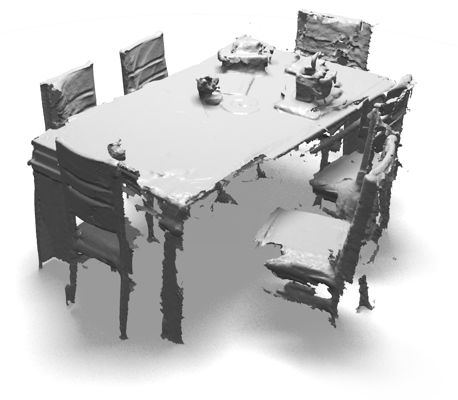} &
			\includegraphics[width = 0.23\textwidth]{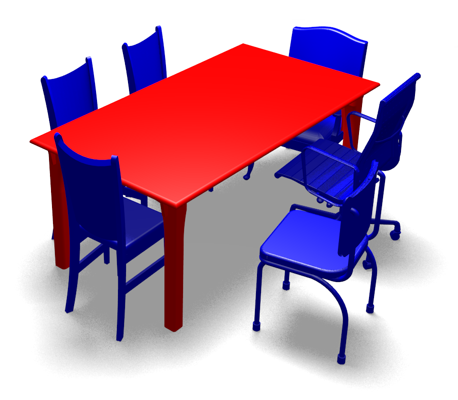} \\
		\end{tabular}		
		\caption{\textbf{Example output of the ASIST algorithm.}  Left: point cloud of a scene acquired using the Kinect sensor.  Right: ASIST output.  Note that semantic invariance has been preserved: chairs replace chairs, and likewise for tables.}
	\end{figure}

	The problem we tackle in this paper is the transformation of 3D scenes.  In particular, we are interested in the subclass of transformations which preserve \emph{semantic invariance}: objects within the scene are to be replaced by other objects from the same class.  Thus, a nightstand should be replaced by another nightstand, and not a packing box.  While a particular packing box may be geometrically similar to the nightstand, it is semantically different and should therefore not be used as the replacement.  Of course, to the extent possible, we would like to preserve geometric similarity as well; the nightstand should ideally be replaced by a nightstand with similar proportions, shape, position, and orientation.  An example is given in Figure \ref{teaser}.
	
	Semantically invariant scene transformation has a number of interesting applications.  In the area of virtual reality, these transformations are useful for designing virtual scenes matching the underlying real scene in which the user is located.  While not critical when the user is stationary, accurate object placement is crucial in any scenario in which the user moves throughout the scene.  For example, semantic invariance means that in sitting on a virtual chair, the user is actually sitting on a real chair.  Beyond preventing injury, this leads to a more realistic VR experience.   In a different application, semantically invariant transformations may be used in the repair of point clouds acquired by the stitching together of many depth images, such as in \cite{izadi2011kinectfusion}.  These fused scans typically have occlusions, holes, and other artifacts, which could be mitigated by the replacement of scene objects with their pristine versions, based on CAD models.  A third application involves mobile robotics systems.  Collision avoidance -- a necessary component of mobile robotics -- is aided by more complete 3D data, and these systems would enjoy benefits similar to the point cloud repair scenario.
	
	While semantically invariant scene transformation is not as standard a problem as, say, object detection or recognition, there have been a few studies in this direction in the last few years.  Both Nan \etal \cite{Nan2012} and Li \etal \cite{Li2015} present approaches to very similar problems, though the approaches themselves differ from ours.  In Nan \etal \cite{Nan2012}, patches of the scene which are likely candidates for replacement are greedily added, and the resulting patch collection is matched against an object database.  Li \etal \cite{Li2015} instead use collections of keypoints, which are then similarly matched against a database.  By contrast, Gupta \etal \cite{Gupta2015} present a complex pipeline rather than a single algorithm.  The pipeline involves many stages, including amongst others: contour detection, perceptual grouping, a convolutional neural network for object detection, instance segmentation, a second convolutional neural network for pose estimation, and registration.  It is also noteworthy that \cite{Gupta2015} is aimed at RGB-D images, rather than point clouds.  Other works which are related, though perhaps not as closely, are SLAM++ of Salas \etal \cite{Salas2013}, and the Sliding Shapes approach of Song and Xiao \cite{song2014sliding}.
	
	\para{Contributions}
	We refer to our approach to the problem as ASIST: Automatic Semantically Invariant Scene Transformation.  ASIST is formulated as a single unified algorithm.  Specifically, both the semantic labeling of the point cloud as well as the replacement of objects within the point cloud are presented as the solution to a single optimization problem.  This contrasts most strongly with the work of Gupta \etal \cite{Gupta2015}, which - while presenting very impressive results - is a complex system composed of many individual algorithms.  We believe that the formulation of a single, unified algorithm for solving semantically invariant scene transformation is an important contribution in its own right.
	
	As has already been noted, another difference from some of the prior work is in ASIST's focus on point clouds, as opposed to RGB-D images.  The differences here are twofold.  First, point clouds are missing the RGB component, which represents a potential advantage for several reasons. Low-power and cheap depth sensors usually do not possess an RGB sensor; moreover, even if they do, it is expensive to keep the RGB sensor on and operating all the time.  Furthermore, the RGB quality is often poor, especially in the most relevant indoor low-light scenario.  Finally, the mechanical limitations of such sensors make the accurate calibration of the RGB and depth cameras a challenging task.  Second, point clouds have more geometric information than depth images.  The acquisition of point clouds has become more common with the growth of systems such as Kinect Fusion \cite{izadi2011kinectfusion} and Project Tango \cite{projectTango}, so focusing on this data structure is a natural choice.
	
	\para{Paper Organization}
	The remainder of the paper is organized as follows.  Section \ref{sec:Related_Work} presents related work in more detail.  Section \ref{sec:Theory} presents the formulation of ASIST, focusing on the unified treatment of semantic labelling and object replacement.  Section \ref{sec:Results} evaluates ASIST on datasets of both synthetic and real point clouds, and comparison with the approaches of Nan \etal \cite{Nan2012} and Li \etal \cite{Li2015}.  Section \ref{sec:Discussion} presents an overall discussion of the algorithm and results, while Section \ref{sec:Conclusions} concludes.
	
	\section{Related Work}
	\label{sec:Related_Work}
	In recent years several studies have explored problems beyond classical object detection or recognition. In Nan \etal\cite{Nan2012} a dataset of deformable models is used to detect object instances in a scanned scene. A given scene is over-segmented into smooth patches, and a RANSAC-like algorithm is used to add patches with high classification scores.  The greedy collection of patches is then compared to models in the database, which are used to remove outliers.
	
	In Salas \etal\cite{Salas2013} object recognition is integrated into a SLAM (simultaneous localization and mapping) pipeline. During the process of integrating new frames into the scanning volume, an existing method~\cite{DrostUNI10} of object detection is employed.  The algorithm as presented is able to search for a relatively small number of models, with limitations primarily due to real-time requirements and the strength of the GPU.
	
	Song \etal\cite{song2014sliding} render a large dataset of models from multiple angles and train a linear SVM classifier for each, according to the exemplar SVM approach of \cite{malisiewicz2011ensemble}. Using a sliding window over a scene, they run each of the classifiers and select a small number of possible models, as well as a best matching pose for each model. Their method produces impressive results at the cost of slow performance.  The sheer number of classifiers and windows to be tested is prohibitive.
	
	Li \etal\cite{Li2015} preprocess a dataset to extract and cluster keypoints from 3D models. Keypoints are arranged into constellations, forming a shape descriptor which is used in real-time 3D scanning to insert model instances into the scene. We use this paper as well as that of Nan \etal\cite{Nan2012}, and the datasets therein, as references for comparison to our method.
	
	Gupta \etal\cite{Gupta2015} use an existing~\cite{guptaECCV14} object detection and segmentation framework and focus on finding object matches in a dataset. They train a convolutional neural network (CNN) to provide a small number of pose and model hypotheses, manually picking five exemplars per category. They then use ICP to perform fine alignment, and a linear classifier to select from the different hypotheses for each object instance.
	
	Mahabadi \etal\cite{karimi2015segment} tackle a related problem: 3D reconstruction of scenes using learned semantic models.  In particular, they introduce a framework which formulates the task of scene reconstruction as a volumetric multi-label segmentation problem. The key idea in their approach is to assign semantic class (including free-space) indicator variables for each voxel.  Inspired by the classical crystallographic technique of Wulff Shapes, they describe how anisotropic surface regularization, which penalizes transitions between labels, can be derived from training data. This naturally continues their line of work from \cite{hane2013joint} and \cite{hane2014class}, in which they propose other alternatives to this binary label penalty term, such as learning preferred directions for inter-label transitions \cite{hane2013joint} or anisotropic regularization for larger (non-convex) object parts \cite{hane2014class}. In contrast to our approach, a unary data term uses information about rays between the camera and each voxel; thus, their work is more applicable to the case of depth images, rather than fused 3D scans as required in our case.  Furthermore, in \cite{karimi2015segment} the semantic labels are assigned to convex model parts and not to whole objects; hence additional computation would be needed to apply this method for the purpose of semantic scene transformation.
	
	Our algorithm uses a dataset of 3D models containing a variety of object classes. We make use of the LightField Descriptor of~\cite{chen2003visual} to find subclasses within the dataset and select a number of exemplars from each category. This descriptor is based on the idea that two similar 3D models look similar from all viewing angles. Multiple orthogonal projections of each model are encoded using Zernike moments and Fourier descriptors. The similarity between two models is then defined as the minimal similarity over the various rotations which can be applied to align the models.
	
	In recent years several datasets and benchmarks for 3D models and scene have emerged. ShapeNet~\cite{shapenet2015} is a collection of CAD models containing 3 million models, of which 220,000 are labeled. ModelNet~\cite{WuSKYZTX15} consists of 600 categories, including 40 main household objects categories.  In the accompanying paper~\cite{WuSKYZTX15}, algorithms for object detection and 2.5D object completion are trained and tested on this dataset. The SunRGBD dataset and benchmark~\cite{Song2015} provides a standard to evaluate and measure the success of scene understanding algorithms. The dataset is densely annotated in both 2D and 3D, using a combination of polygons and bounding boxes.
	
	\section{Automatic Semantically Invariant Scene Transformation}
	\label{sec:Theory}
	\subsection{General Approach}
	
	Our goal is to take a point cloud representing a real world scene as input, and to transform it while remaining faithful to the semantics of the scene.  More specifically, given a fixed set of object classes and a database of objects from these classes, we wish to:
	\begin{enumerate}
		\item recognize instances of objects within the scene belonging to these classes;
		\item replace these instances with geometrically similar objects of the same class from the database, respecting original positions and orientations.
	\end{enumerate}
	The ideal output of the algorithm would then be a semantically similar scene, but with new objects placed within the scene.  The potential applications of such a technique have been highlighted in Section \ref{sec:Introduction}.
	
	In order to arrive at a more unified treatment of the problem, which also yields superior results, we treat both problems \emph{simultaneously}.  That is, our goal is to solve both the semantic segmentation and replacement problems at the same time and in a consistent manner.  In what follows, we show how to formulate the problem to this end.

	\subsection{Cell Classification}
	\label{sec:Cell_Classification}
	
	We are given a point cloud representing the scene.  We voxelize the scene, and organize these voxels into larger structures called \emph{cells}.  A cell is defined to be a cubic collection of voxels; that is, it is a patch of $m \times m \times m$ voxels.  In practice, we take $m = 9$.  Note that much of the scene is empty, so it may appear that voxelization is a wasteful process; however, we only perform computations on occupied cells, i.e.~cells containing points from the scene's point cloud.
	
	Each voxel is regarded as the center of a cell surrounding it.  We train a random forest \cite{criminisi2012decision} which classifies cells according to one of $n_c$ fixed classes, or ``clutter'', to which class label $0$ is assigned.  The forest performs its classification using split functions which are a generalization of decision stumps and binary decisions. More specifically, we compute a \emph{base feature} for each voxel within the cell; we use three types of base features, namely binary occupancy, the distance function of the voxel's center from the point cloud, and the height of the cell as measured by the height of its center voxel (for further implementation details, refer to Section \ref{sec:Implementation}).  Thus, the base features may be written as $\mbr{h}_k \in \mathbb{R}^{M_k}$; in the case of the binary occupancy and distance function, $M_k = m^3$, whereas in the case of the height $M_k = 1$.  The forest's split functions are then represented as triples $(k, \mbr{u}, \tau)$ where $k \in \{1, \dots, K\}$, $\mbr{u} \in \mathbb{R}^{M_k}$, and $\tau \in \mathbb{R}$; the decision is then based on the value of the binary variable $\indicator[\mbr{u}^T \mbr{h}_k > \tau]$, where $\indicator[\cdot]$ denotes the indicator function.  A decision stump corresponds to $\mbr{u}$ with only a single non-zero value; a pairwise decision has two non-zero values; and so on.  We describe the details of the feature choices we use in practice in Section \ref{sec:Implementation}.
	
	This procedure yields forest scores for each voxel; we then assign forest scores to each point $p$ in the point cloud $\mathcal{P}$ via nearest neighbors (though any simple interpolation scheme will do).  Thus, at the end of this process we have a collection of forest scores $f_{cp}$, where $f_{cp}$ indicates the score of class $c$ for point $p$.  As is customary for random forests, the scores for a given point form a probability distribution, i.e. $f_{cp} \ge 0$ and $\sum_c f_{cp} = 1$.

	\subsection{Joint Semantic Segmentation and Object Replacement}
	\label{sec:Joint_Semantic}
	
	We are given a dataset of objects which come from the $n_c$ fixed classes.  To reduce computational complexity, the objects from each class within the database are clustered into a set of groups, each of which is represented by an \emph{exemplar}.  We achieve this by first doing a rough clustering based on scale using $k$-medoids; within each new cluster, we then further sub-cluster based on the LightField Descriptor \cite{chen2003visual}, again using k-medoids.  At the end of the process, we have a collection $\mathcal{E}$ of exemplars taken from all classes.
	
	Our goal is now to decide which of these exemplars to insert into our scene, where to insert them, and in which pose.  To this end, we define the variables $w_{ep}$, representing the weight of an exemplar $e$ at point $p$.  For a fixed point, these weights can be thought of as a probability distribution over which exemplar should be inserted at that point.  Ideally, then, we would have $w_{ep} = 1$ for exactly one exemplar $e$, and is $w_{ep} = 0$ for the rest.  In practice, we will simply require the probability distribution condition, that is
	\[
	\sum_e w_{ep} = 1 \quad \text{and} \quad w_{ep} \ge 0
	\]
	for every $e$ and $p$.  By convention, $e = 0$ corresponds to ``clutter'', or non-object; that is, $w_{0p} = 1$ is an indication not to insert any object at point $p$.  Note that this representation is similar to the one used in \cite{LitanyBB12}.
	
	Properly choosing the weights $w_{ep}$ effectively solves a soft version of the semantic segmentation problem: we can assign a soft/probabilistic label to each point in the scene according to which exemplar that point corresponds to.  But in addition, we wish to solve the object replacement problem.  Replacement entails deciding which exemplars to insert into the scene, and in which pose -- comprising both position and orientation.  Neither the identity of the exemplars to be included nor their pose follows directly from the soft weights $w_{ep}$.
	
	To solve the object replacement problem, therefore, we add two new sets of variables. (1) The variables $v_e \in [0,1]$ are ``votes'' for the exemplar $e$, and indicate whether an exemplar should be inserted.  A positive vote $v_e > 0$ indicates the exemplar is to be inserted in the scene, while a vote of $v_e = 0$ implies it should be left out. (2) The transformations $T_e$ denote the pose in which a candidate exemplar should be inserted within the scene.  We take $T_e$ to belong to the set of rigid transformations (translations and rotations), though one could broaden this to including scaling or more exotic non-rigid transformations.
	
	We propose to perform semantic segmentation and object replacement jointly, by minimizing an energy which is the sum of six terms, that is:
	\[
	\mr{E}\left(\{w_{ep}\}, \{T_e\}, \{v_e\}\right) = \sum_{i=1}^6 \lambda_i \mr{E}_i\left(\{w_{ep}\}, \{T_e\}, \{v_e\}\right)
	\]
	The six individual energy terms are as follows:
	\begin{align*}
		& \mr{E}_1 = \sum_{p, c} \left( f_{cp} - \sum_e A_{ce} w_{ep} \right)^2 \\
		& \mr{E}_2 = \sum_{p, e} w_{ep} D(x_p, T_e \mathcal{X}_e)  \\
		& \mr{E}_3 = \sum_e \sum_{p_1, p_2} L_{p_1p_2} w_{ep_1} w_{ep_2} \\
		& \mr{E}_4 = \sum_p \sum_e |w_{ep}|^\ell \\
		& \mr{E}_5 = -\sum_e \left(\sum_p w_{ep}\right) v_e  \\
		& \mr{E}_6 = \sum_{e1, e2} \mr{Q}(T_{e_1} \mathcal{X}_{e_1} \, , T_{e_2} \mathcal{X}_{e_2}) \, v_{e_1} v_{e_2}
	\end{align*}
	The minimization is performed subject to the following constraints:
	\[
	\sum_e w_{ep} = 1 \,\, \forall p \quad \quad w_{ep} \ge 0 \,\, \forall p, e \quad \quad 0 \le v_e \le 1 \,\, \forall e
	\]
	Let us take each of these terms in turn.
	\begin{itemize}
		\item \itemtitle{$\mr{E}_1$ -- Semantic Data Term.}  The constants $A_{ce}$ are defined as follows: $A_{ce} = 1$ if exemplar $e$ belong to class $c$ and $0$ otherwise, reflecting the assignment of exemplars to classes known a priori.  Thus, the semantic data term tries to drive the weights to be faithful to the forest classifier; the sum of the weights over all of the exemplars in a particular class should match the output of the forest classifier for that class.
		\item \itemtitle{$\mr{E}_2$ -- Geometric Data Term.}  $x_p$ is the location of the point $p$, and $\mathcal{X}_e$ is the point cloud which represents exemplar $e$. $T_e \mathcal{X}_e$ indicates the point cloud resulting from applying transformation $T_e$ to each point in $\mathcal{X}_e$.  Finally, $D$ denotes an extrinsic distance between two sets of points.  Thus, the geometric data term is a kind of weighted distance term.  The goal is to match the scene points as best as possible to a given exemplar.  Note that this term only pays attention to exemplars with non-zero weights.  For the clutter exemplar $e=0$, $D(x_p, T_e \mathcal{X}_e)$ is not well-defined; we replace it by a constant $D_{clutter}$ for all points $p$. The actual value of $D_{clutter}$ used is discussed in Section \ref{sec:Results}.
		\item \itemtitle{$\mr{E}_3$ -- Spatial Smoothness Term.}  This term is a spatial regularization on the weights.  In particular, we strive to smooth out the weights considered as a function of spatial location; that is, for each exemplar $e$, we wish to smooth out $w_{ep}$, considered as a function of $p$.  We achieve this by using a Laplacian smoothing term; here $\mbr{L}$ is the Laplacian matrix over a weighted graph defined over the point cloud.  More details are given in Section \ref{sec:Implementation}.
		\item \itemtitle{$\mr{E}_4$ -- Sparsity Term.}  The parameter $\ell$ is chosen to be in the range $(0,1)$.  Thus, this term promotes sparsity on the weights.  Recalling that for each pixel $p$ the weights $w_{ep}$ are normalized to sum to 1, this term encourages all of the mass to be placed on a single exemplar.  In practice, we choose $\ell = 0.1$.
		\item \itemtitle{$\mr{E}_5$ -- Weight-Vote Agreement Term.}  This term prefers to choose a large vote $v_e$ for exemplar $e$ when the sum of the weights for that exemplar, over all pixels, is large. 
		\item \itemtitle{$\mr{E}_6$ -- Non-Collision Term.}  $\mr{Q}(\mathcal{X}_1, \mathcal{X}_2)$ denotes the binary overlap between two shapes $\mathcal{X}_1$ and $\mathcal{X}_2$ i.e. $\mr{Q}(\mathcal{X}_1, \mathcal{X}_2) = 1$ if $\mathcal{X}_1$ and $\mathcal{X}_2$ overlap and $0$ otherwise.  Thus $\mr{Q}(T_{e_1} \mathcal{X}_{e_1} \, , T_{e_2} \mathcal{X}_{e_2})$ measures the binary overlap between the exemplars $e_1$ and $e_2$, in the poses $T_{e_1}$ and $T_{e_2}$, respectively.  Given that the votes $v_e$ are constrained to be non-negative, this term encourages us to select only non-overlapping exemplars.
	\end{itemize}
	Having thus defined an energy which captures the idea of joint semantic segmentation and object replacement, we proceed to describe how to minimize the energy.

	\subsection{Simplifying the Energy}
	
	We begin by rewriting the energy more neatly in matrix-vector notation.  Suppose that the number of points, exemplars, and classes are $n_p$, $n_e$, $n_c$, respectively.  We form the vector $\mbr{w} \in \mathbb{R}^{n_e n_p}$ by stacking the weights, as
	\[
	\mbr{w} = [w_{11}, \dots, w_{1 n_p}, \dots, w_{n_e 1}, \dots, w_{n_e n_p}]^T.
	\]
	We similarly stack the outputs of the random forest into a vector $\mbr{f} \in \mathbb{R}^{n_c n_p}$ as
	\[
	\mbr{f} = [f_{11}, \dots, f_{1 n_p}, \dots, f_{n_c 1}, \dots, f_{n_c n_p}]^T.
	\]
	We denote $d_{ep}(T_e) = D(x_p,T_e\mathcal{X}_e)$, and form the vector of distances $\mbr{d}(\mbr{T}) \in \mathbb{R}^{n_e n_p}$ by
	\[
	\mbr{d}(\mbr{T}) = [d_{11}(T_1), \dots, d_{1 n_p}(T_1), \dots, d_{n_e 1}(T_{n_e}), \dots, d_{n_e n_p}(T_{n_e})]^T.
	\]
	We keep the transformation ($\mbr{T}$) dependence explicit as we will be optimizing over the transformations.  Finally, we define $n_p$ matrices $\mbr{R}_p \in \mathbb{R}^{n_p \times n_e n_p}$, which pick out the entries of $\mbr{w}$ related only to point $p$.  In other words,
	\[
	\mbr{R}_p \mbr{w} = [w_{1 p}, \dots, w_{n_e p}]^T
	\]
	We define a similar set of sampling matrices for the forest entries $\mbr{f}$, which we denote $\mbr{S}_p$.  Finally, we denote by $\mbr{Q}(\mbr{T})$ the $n_e \times n_e$ matrix with entries $\mr{Q}(T_{e_1} \mathcal{X}_{e_1} \, , T_{e_2} \mathcal{X}_{e_2})$.
	
	We may then rewrite the terms of the objective as
	\begin{align*}
		& \mr{E}_1 = \mbr{w}^T \left(\sum_p \mbr{R}_p^T \mbr{A}^T \mbr{A} \mbr{R}_p \right) \mbr{w}  -2\mbr{f}^T \left(\sum_p \mbr{S}_p^T \mbr{A} \mbr{R}_p \right) \mbr{w} \\
		& \mr{E}_2 = \mbr{d}(\mbr{T})^T\mbr{w}  \\
		& \mr{E}_3 = \mbr{w}^T (\mbr{I}_{n_e} \otimes \mbr{L}) \mbr{w} \\
		& \mr{E}_4 = -\|\mbr{w}\|_\ell^\ell \\
		& \mr{E}_5 = -\mbr{v}^T (\mbr{I}_{n_e} \otimes \mbr{1}_{n_p}^T) \mbr{w} \\
		& \mr{E}_6 = \mbr{v}^T\mbr{Q}(\mbr{T})\mbr{v} \\
	\end{align*}
	where $\otimes$ is the Kronecker product; $\mbr{I}_k$ is the identity matrix of size $k \times k$; and $\mbr{1}_k$ is the vector whose entries are all $1$, of dimension $k$.  Simplifying, we then have
	\[
	\mr{E} = \mbr{w}^T \mbg{\Psi}_{ww} \mbr{w} + \mbg{\theta}_w(\mbr{T})^T \mbr{w} + \xi_w \|\mbr{w}\|_\ell^\ell + \mbr{v}^T \mbg{\Psi}_{vv}(\mbr{T}) \mbr{v} + \mbr{v}^T \mbg{\Psi}_{vw} \mbr{w}
	\]
	where
	\begin{align*}
		& \mbg{\Psi}_{ww} = \lambda_1 \sum_p \mbr{R}_p^T \mbr{A}^T \mbr{A} \mbr{R}_p + \lambda_3 \mbr{I}_{n_e} \otimes \mbr{L} \\
		& \mbg{\theta}_w(\mbr{T}) = -2 \lambda_1 \left(\sum_p \mbr{R}_p^T \mbr{A}^T \mbr{S}_p \right) \mbr{f} + \lambda_2 \mbr{d}(\mbr{T}) \\
		& \xi_w = -\lambda_4 \\
		& \mbg{\Psi}_{vv}(\mbr{T}) = \lambda_6 \mbr{Q}(\mbr{T}) \\
		& \mbg{\Psi}_{vw} = -\lambda_5 \mbr{I}_{n_e} \otimes \mbr{1}_{n_p}^T
	\end{align*}
	The constraints may be simplified as follows:
	\[
	\mbg{\Gamma} \mbr{w} = \mbr{1} \quad \quad \quad \mbr{w} \ge \mbr{0} \quad \quad \quad \mbr{0} \le \mbr{v} \le \mbr{1}
	\]
	where $\mbg{\Gamma}$ is a $n_p \times n_en_p$ matrix, whose $p^{th}$ row is equal to $\mbr{1}_{n_p}^T \mbr{R}_p$.

	\subsection{Minimizing the Energy}
	
	Our problem is now
	\[
	\min_{\mbr{T},\mbr{w},\mbr{v}} \mr{E} = \mbr{w}^T \mbg{\Psi}_{ww} \mbr{w} + \mbg{\theta}_w(\mbr{T})^T \mbr{w} + \xi_w \|\mbr{w}\|_\ell^\ell + \mbr{v}^T \mbg{\Psi}_{vv}(\mbr{T}) \mbr{v} + \mbr{v}^T \mbg{\Psi}_{vw} \mbr{w}
	\]
	subject to
	\[
	\mbg{\Gamma} \mbr{w} = \mbr{1} \quad \quad \quad \mbr{w} \ge \mbr{0} \quad \quad \quad \mbr{0} \le \mbr{v} \le \mbr{1}
	\]
	We use a technique based on alternating minimization.  In particular, we cycle through subproblems with respect to $\mbr{T}$, $\mbr{w}$, and $\mbr{v}$.  We now detail how to solve these individual subproblems.
	
	\para{Minimization \wrt $\mbr{T}$}
	Fixing $\mbr{w}$ and $\mbr{v}$, the minimization \wrt $\mbr{T}$ reduces to
	\[
	\min_{\mbr{T}} \mbg{\theta}_w(\mbr{T})^T \mbr{w} + \mbr{v}^T \mbg{\Psi}_{vv}(\mbr{T}) \mbr{v}
	\]
	There are two conditions under which the second term, $\mbr{v}^T \mbg{\Psi}_{vv}(\mbr{T}) \mbr{v}$ is equal to $0$.  The first condition is that the coefficient on the non-collision term $\lambda_6 = 0$.  The second condition is that positive overlap between two exemplars implies that at most one is selected; that is, $\mr{Q}(T_{e_1} \mathcal{X}_{e_1} \, , T_{e_2} \mathcal{X}_{e_2}) > 0 \Rightarrow v_{e_1} = 0 \text{ or } v_{e_2}=0$.  As we shall see, in practice one of these two conditions generally holds.  That is, as we describe in both Algorithm \ref{alg:main} and Section \ref{sec:Implementation}, for the initial iteration $\lambda_6$ is set to $0$, so that the first condition is satisfied.  In later iterations when $\lambda_6 > 0$, we observe empirically that the second condition generally holds, at least approximately.  We use this as justification to ignore the second term $\mbr{v}^T \mbg{\Psi}_{vv}(\mbr{T}) \mbr{v}$.
	
	In this case, the optimization problem reduces to
	\[
	\min_{\mbr{T}} \mbg{\theta}_w(\mbr{T})^T \mbr{w}
	\]
	The role of the transformations $\mbr{T}$ is somewhat obscured within the matrix-vector formulation.  Going back to the original formulation, we can rewrite the above as
	\[
	\min_{T_1, \dots, T_{n_e}} = \sum_{p, e} w_{ep} D(x_p, T_e \mathcal{X}_e)
	\]
	It is easy to see that this minimization problem is separable, i.e.~it decomposes into a separate minimization for each exemplar $e$.  So for each $e$ one needs to solve
	\begin{equation}
		\min_{T_e} \sum_p w_{ep} D(x_p, T_e \mathcal{X}_e)
		\label{eq:min_T}
	\end{equation}
	Solving (\ref{eq:min_T}) for the optimal transformation $T_e$ is the same as solving a weighted ICP problem.  This can be done with standard techniques, e.g.~\cite{rusinkiewicz2001efficient}.
	
	\para{Minimization \wrt $\mbr{w}$}
	The main issue here is the sparsity term $\|\mbr{w}\|_\ell^\ell$, which is not convex.  However, since we are already performing an iterative optimization, it is natural to use the iterative reweighted least squares (IRLS) technique.  This allows us to replace $\|\mbr{w}\|_\ell^\ell$ with 
	\[
	\sum_e \eta_{ep} w_{ep}^2
	\]
	where 
	\begin{equation}
		\eta_{ep} = \left|w_{ep}^{(k-1)}\right|^{\ell-2}
		\label{eq:eta_update}
	\end{equation}
	and $w_{ep}^{(k-1)}$ are the optimal weights from the previous, i.e. $(k-1)^{th}$ iteration.  (For example, see \cite{chartrand2008iteratively}.)  In this case, the energy becomes
	\[
	\tilde{\mr{E}} = \mbr{w}^T \tilde{\mbg{\Psi}}_{ww} \mbr{w} + \mbg{\theta}_w(\mbr{T})^T \mbr{w} + \mbr{v}^T \mbg{\Psi}_{vv}(\mbr{T}) \mbr{v} + \mbr{v}^T \mbg{\Psi}_{vw} \mbr{w}
	\]
	where
	\[
	\tilde{\mbg{\Psi}}_{ww} = \mbg{\Psi}_{ww} + \diag(\mbg{\eta})
	\]
	Now, fixing $\mbr{v}$ and $\mbr{T}$, it is clear that the minimization of $\tilde{\mr{E}}$ \wrt $\mbr{w}$ is a convex quadratic program, i.e.
	\begin{equation}
		\min_\mbr{w} \mbr{w}^T \tilde{\mbg{\Psi}}_{ww} \mbr{w} + \left(\mbg{\theta}_w(\mbr{T}) + \mbg{\Psi}_{vw}^T \mbr{v}\right)^T \mbr{w} \quad \text{s.t.} \quad \mbg{\Gamma} \mbr{w} = \mbr{1}, \quad \mbr{w} \ge \mbr{0}
		\label{eq:min_w}
	\end{equation}
	Thus, one can solve this step for the global minimum \wrt $\mbr{w}$ (\emph{not} the global minimum of the entire function, just of this step) using standard solvers.
	
	\para{Minimization \wrt $\mbr{v}$}
	One can easily observe that in fixing $\mbr{w}$ and $\mbr{T}$, the minimization \wrt $\mbr{v}$ is a quadratic program, that is
	\begin{equation}
		\min_\mbr{v} \mbr{v}^T \mbg{\Psi}_{vv}(\mbr{T}) \mbr{v} + \left(\mbg{\Psi}_{vw} \mbr{w}\right)^T \mbr{v} \quad \text{s.t.} \quad \mbr{0} \le \mbr{v} \le \mbr{1}
		\label{eq:min_v}
	\end{equation}
	However, given that $\mbr{Q}(\mbr{T})$, the collision matrix, is not positive semidefinite, the QP will not, in general, be convex. Thus, while one may use standard solvers, this step will not in general yield the global minimum \wrt $\mbr{v}$; rather, a local minimum is all that can be guaranteed.

	\para{The Algorithm} The overall algorithm is summarized in Algorithm \ref{alg:main}.  The main structure of the iterations consists of three nested loops: the outer loop over $i$, in which the energy function's coefficients are adjusted; the middle loop over $j$, which contains the minimizations over $\mbr{T}$ (registration step), and $\mbr{v}$ (voting step); and an inner loop over $k$, which allows for the IRLS iterations necessary for minimization over $\mbr{w}$ (segmentation step).
	
	There are several details in the initialization which have not yet been covered.  We now proceed to discuss these and other implementation details.
	
	\begin{algorithm}[t]
		\fbox{
			\parbox{0.45\textwidth}{
				\KwIn{point cloud $\mathcal{P}$, set of exemplars $\mathcal{E}$}	
				\KwOut{set of exemplars $\mathcal{E}_{rep} \subset \mathcal{E}$ to insert into the scene, pose $T_e$ for each exemplar $e \in \mathcal{E}_{rep}$}
				\BlankLine
				\emph{Initialization:}
				\begin{itemize}
					\item \nvs set the sequence of coefficients $\left\{\lambda_6^{(i)}\right\}_{i=1}^{N_{out}}$
					\item \nvs evaluate Random Forest to get initial confidence per class for each point $f_{cp}$
					\item \nvs run Mean Shift to set initial exemplar positions
					\item \nvs set initial $\mbr{w}$ according to forest scores: \\ $w_{ep} \leftarrow f_{cp} \, / \, (\text{\# exemplars in class }c)$
					\item \nvs at every position run $N_{ICP}$ weighted ICP's with different initial rotation around $z$-axis; keep result with smallest distance
					\item \nvs set initial $\mbr{v} \leftarrow \mbr{1}$
				\end{itemize}
				\emph{Iterative Minimization:}
				\BlankLine
				
				\For{$i\leftarrow 1$ \KwTo $N_{out}$}{
					$E^{(i)} \leftarrow$ energy function with coefficients $\lambda_6 = \lambda_6^{(i)}$ \\
					\For{$j\leftarrow 1$ \KwTo $N_{in}$}{
						registration step -- compute $\mbr{T}$: solve (\ref{eq:min_T}) \\
						\For {$k\leftarrow 1$ \KwTo $N_{IRLS}$}{
							set $\mbg{\eta}$ according to (\ref{eq:eta_update}) \\
							segmentation step -- compute $\mbr{w}$: solve (\ref{eq:min_w})
						}
					}
					voting step -- compute $\mbr{v}$: solve (\ref{eq:min_v})
				}
				\BlankLine
				$\mathcal{E}_{rep} \leftarrow \{ e \in \mathcal{E}: \,\, v_e > 0, \,\,\, \sum_p w_{ep} \ge \text{threshold} \}$
			}}
			\BlankLine
			\BlankLine
			\caption{ASIST algorithm. }
			\label{alg:main}
		\end{algorithm}

	\subsection{Implementation Details}
	\label{sec:Implementation}
	In this section we explain in details the algorithm pipeline.  Please refer to Algorithm \ref{alg:main} for each of the relevant steps.
	
	\para{Sequence of Energy Coefficients}   
	The energy function is, in general, non-convex in $\mbr{v}$; this is due to the indefinite nature of the collision matrix $\mbr{Q}(\mbr{T})$, and hence of the matrix $\mbg{\Psi}_{vv}(\mbr{T}) = \lambda_6\mbr{Q}(\mbr{T})$.
	
	To alleviate this non-convexity, we define a monotonically increasing sequence of coefficients $\{\lambda_6^{(i)}\}_{i=1}^{N_{out}}$ on the non-collision term $\mr{E}_6$, and corresponding sequence of energy functions $\{\mr{E}^{(i)}\}_{i=1}^{N_{out}}$. Choosing $\lambda_6^{(1)}=0$ we get an initial energy function which is convex in $\mbr{v}$. Using a small non-collision coefficient $\lambda_6$ will generally result in weights $w_{ep}$ which are ``non-decisive'', in that we will not have $w_{ep} = 1$ for a single $e$; rather, for a given point several exemplars will have positive weights.  By gradually increasing the non-collision coefficient $\lambda_6$, the decisiveness of the weights improves at the cost of increasing non-convexity.
	
	We observe experimentally that using a moderate growth rate for the non-collision coefficients $\lambda_6$ increases the probability of converging to the correct results.  This is further ameliorated by initializing the quadratic program with the solution for $\mbr{v}$ from the previous iteration.
	
	\para{Random Forest}
	Recall that in Section \ref{sec:Cell_Classification}, we described the computation of $K$ base features per cell, where each such base feature is a vector $\mbr{h}_k \in \mathbb{R}^{M_k}$.  There are two separate types of base features that are used:
	\begin{enumerate}
		\item Scalar field features.  These include the occupancy and distance function features.  In this case, the size of the feature $M_k = m^3$, where the cell is $m \times m \times m$.
		\item Scalar features.  We used only such feature, the height of the cell, as measured by the height of the center voxel of the cell.  In this case, the size of the feature is $M_k = 1$.
	\end{enumerate}
	In describing the split functions used, we will treat scalar field features as functions in three dimensions, i.e. $\mr{h}_k(x,y,z)$ over all voxels $(x, y, z)$ in the cell, with the understanding that we can convert from functional notation $\mr{h}_k(\cdot)$ to the vector notation $\mbr{h}_k$ used in Section \ref{sec:Cell_Classification} simply by stacking.  Thus, whereas the split functions were previously described as $\omega(k, \mbr{u}, \tau) = \indicator[\mbr{u}^T \mbr{h}_k > \tau]$, for the sake of a simpler explanation, we will now describe those corresponding to scalar field features by
	\[
	\omega(k, \mr{u}, \tau) = \indicator{}\left[\sum_{(x, y, z) \in \text{Cell}} \mr{u}(x, y, z) \,\, \mr{h}_k(x, y, z) > \tau\right].
	\]
	
	In all experiments we obtained a prior on the label probability of each point in the point cloud using a random forest classifier trained to maximize the Shannon Entropy at each split. The split functions $\omega$ used in our experiments were designed so that some enable rotation invariant splits, while others offer more rotation selective ones.  In what follows, the $z$-direction is the vertical direction, measuring height off the ground.
	
	\begin{itemize}
		\item \itemtitle{Height:} Let $h_3$ be the height of the center voxel of the cell.  Then $\omega = \indicator[h_3 > \tau]$.
		\item \itemtitle{Rotation Invariant:} Uses a dot product between a radially symmetric weight vector and the cells feature vector. $\omega = \indicator[\sum_{(x,y,z)\in \text{Cell}} \mr{u}(x,y,z)\mr{h}_k(x,y,z)>\tau]$ where $\mr{u}$ is a randomly generated function with values that are rotationally symmetric for rotations around the $z$-axis.
		\item \itemtitle{Box:} Sums the occupancy or the distance function in a randomly generated box.  Thus, $\omega = \indicator[\sum_{(x,y,z)\in B}\mr{h}_k(x,y,z)>\tau]$ where $B$ is a box with a random location and size within the relevant cell, and $\mr{h}_k(x,y,z)$ is either the occupancy or distance function value of the voxel at location $(x,y,z)$.
		\item \itemtitle{Horizontal Slab:} Sums the occupancy or distance function over all $9\times9$ voxels at a single $z$-value.  The slab can be written as $R(z_0) = \{(x,y,z)\in \text{Cell}: z = z_0\}$.   Then the split function is $\omega = \indicator[\sum_{(x,y,z)\in R(z_0)}\mr{h}_k(x,y,z)>\tau]$, where $z_0 \in\{-4,-3,...,4\}$ is selected randomly.
		\item \itemtitle{Pixelwise Values:} Linear combination of the occupancy or the distance function values of up to three voxels  $\omega = \indicator[\sum_{i=1}^3{a_i\mr{h}_k(x(v_i),y(v_i),z(v_i))}>\tau]$ where the voxels $v_i$ are selected randomly within the cell limit and $a_i\in\{-1,0,1\}$ is also selected at random.
	\end{itemize}
	
	For all split functions the value of $\tau$ is selected randomly within the feasible bounds of the training data reaching the node.
	
	\para{Mean Shift for Initial Exemplar Positions} 
	Initial exemplar locations are obtained by running a fast version of a weighted Mean Shift \cite{fukunaga1975estimation} algorithm on the point cloud after is has been projected onto the $xy$-plane.  The algorithm is run once for each object class; during the run for object class $c$, the weight assigned to point $p$ is taken to be $f_{cp}$, the random forest output for that class.  The modes found by Mean Shift therefore tend to be at locations where the forest assigned high probabilities to the class in question.
	
	The bandwidth of the Mean Shift algorithm is set differently for each class, based on a rough estimate of the object size in that class. As a post-processing step, we merge modes that are closer than $1.5$ times the bandwidth.  The modes returned for a given class are then assigned as the starting positions for each exemplar representing that class.
	
	The next step in the initialization process is to determine the starting orientation of each exemplar. This is done by initializing each exemplar in $8$ equally spaced rotations around the $z$-axis, running an ICP algorithm to finely tune the position and orientation of the exemplar, and then choosing the best fitted exemplar out of the $8$. In cases in which the best exemplar is farther than a predetermined threshold all $8$ candidates are removed from consideration.
	
	\para{Laplacian Operator $\mbr{L}$ for Spatial Smoothness Term $\mr{E}_3$} 
	Given the point cloud $\mathcal{P}$ we construct an undirected weighted graph $G(\mathcal{P},\mathcal{F}, \mbg{\Omega})$, where the edge set $\mathcal{F}$ is constructed using $k$-nearest neighbors.  The edge weights are given by $\Omega_{pq} = e^{-\|x_p-x_q\|^2/2\sigma^2}$ for $(p,q) \in \mathcal{F}$ and are zero otherwise.  We use the random walk version of the normalized graph Laplacian of $G$ \cite{chung1997spectral} where $\mbr{L} = \mbr{I}-\mbg{\Delta}^{-1}\mbg{\Omega}$, and $\mbg{\Delta} = \diag(\sum_q \Omega_{pq})$. In all of our experiments we used $k=10$ nearest neighbors and $\sigma = 5$.
	
	\para{Output Exemplar Filtering} 
	Recall from Section \ref{sec:Joint_Semantic} that our proposed criterion for inserting exemplars was based entirely on their vote: if $v_e > 0$ exemplar $e$ would be inserted, and otherwise it would not.  In practice, this works quite well most of the time, as the algorithm generally produces a set of non-overlapping exemplars.
	
	However, it may sometimes be the case that an exemplar is assigned in error to a small number of isolated points. This rarely happens since such points are usually classified as clutter by the forest, and thus Mean Shift finds no corresponding mode.  But in the unusual case that a mode is found, then even if the vote $v_e$ corresponding to that mode is small, it is still positive.  Thus, the original criterion requires that we retain this exemplar, which is problematic.
	
	To handle such cases we apply a straightforward filter as a final step, where we keep only exemplars $e$ for which the vote $v_e$ is positive, and at the same time the aggregation over the weights of the exemplar $\sum_p w_{ep}$ exceeds a small predetermined threshold.  We set the threshold to be $0.1$ in all our experiments.

	\subsection{Speeding Up the Algorithm}
	
	There are three steps to our algorithm, corresponding to the three sets of variables we minimize with respect to.  Each of the these steps must be repeated several times, as shown in Algorithm \ref{alg:main}.
	
	Minimizing with respect to the transformations $T_e$ does not pose a speed problem, as weighted ICP may be solved quickly (see for example \cite{spagnuolo2012parallelized}).  The issue is the two quadratic programs we must solve, one with respect to $\mbr{w}$ and the other with respect to $\mbr{v}$.  However, these two QPs are quite different in scale.  The QP with respect to $\mbr{w}$ is of size $n_e n_p$, while the QP with respect to $\mbr{v}$ is of size $n_e$.  Rough orders of magnitude for these two variables are $10^5-10^6$ for $n_p$ and $10^2$ for $n_e$.  Thus, the QP for $\mbr{v}$ can be solved with a standard solver quite quickly; while the QP for $\mbr{w}$, which will be of size $n_e n_p \approx 10^7 - 10^8$, is considerably slower to solve in a standard fashion.  We thus propose two independent techniques to speed up this QP.
	
	\para{Subspace Parameterization}
	Let us denote by $\mbr{w}_e \in \mathbb{R}^{n_p}$ the vector $[w_{e,1}, \dots, w_{e,n_p}]^T$.  This is the vector of weights corresponding to just the exemplar $e$, over all points.  We may therefore think of $\mbr{w}_e$ as a scalar field or function on the volume.
	
	The idea is to represent this scalar function in a more parsimonious fashion.  Thus, we use an expansion in terms of basis functions.  A natural set of basis functions is provided by the Laplacian operator $\mbr{L}$, which we already use in the smoothing term.  We take the basis functions to the eigenfunctions of $\mbr{L}$, corresponding to the smallest $n_b$ eigenvalues.  (Recall that the smaller the eigenvalue of the Laplacian, the smoother the function.)  We denote these functions by $\{\mbg{\phi}_i\}_{i=1}^{n_b}$, where each $\mbg{\phi}_i \in \mathbb{R}^{n_p}$, and the collection is given by the matrix whose columns are the individual functions, i.e. $\mbg{\Phi} \in \mathbb{R}^{n_p \times n_b}$.  In practice, we find $n_b = 30$ Laplacian basis functions suffice.  
	
	In this case, we can represent the function $\mbr{w}_e$ as $\mbg{\Phi} \mbg{\alpha}_e$, for $\mbg{\alpha}_e \in \mathbb{R}^{n_b}$. Stacking the coefficients of each exemplar to get a vector $\mbg{\alpha} \in \mathbb{R}^{n_b n_e}$, we have that
	\[
	\mbr{w} = (\mbr{I}_{n_e} \otimes \mbg{\Phi}) \mbg{\alpha} \equiv \hat{\mbg{\Phi}} \mbg{\alpha}
	\]
	And thus, we may represent the entire vector $\mbr{w}$ with only $n_b n_e$ values.  Given that $n_b = 30$ in practice, this is a huge complexity reduction of 3-4 orders of magnitude.
	
	The energy we need to minimize thus becomes
	\begin{align*}
		\tilde{\mr{E}} & = \mbr{w}^T\tilde{\mbg{\Psi}}_{ww}\mbr{w} + (\mbg{\theta}_w(\mbr{T}) - \mbg{\Psi}_{vv}(\mbr{T}) \mbr{v})^T \mbr{w} \\
		& = \mbg{\alpha}^T (\hat{\mbg{\Phi}}^T \tilde{\mbg{\Psi}}_{ww} \hat{\mbg{\Phi}}) \mbg{\alpha} + [(\mbg{\theta}_w(\mbr{T}) - \mbg{\Psi}_{vv}(\mbr{T}) \mbr{v})^T \hat{\mbg{\Phi}}] \mbg{\alpha} \\
		& \equiv \mbg{\alpha}^T \mbg{\Psi}_{\alpha\alpha} \mbg{\alpha} + \mbg{\theta}_\alpha(\mbr{T})^T \mbg{\alpha}
	\end{align*}
	The two set of constraints are now represented in terms of the coefficients $\mbg{\alpha}$ as
	\[
	\mbg{\Gamma} \hat{\mbg{\Phi}} \mbg{\alpha} = \mbr{1} \quad \quad \quad \hat{\mbg{\Phi}} \alpha \ge \mbr{0}
	\]
	It is not clear if the equality constraints are even still feasible, given that we are dealing with a much smaller number of variables.  It turns out they are still feasible, which we now show.  To do so, it will be easier to go back to the original non-vectorized formulation.  These constraints were $\sum_e w_{ep} = 1$ for all $p$.  But this means that
	\[
	\sum_e \mbr{w}_e = \mbr{1}_{n_p} \Rightarrow \sum_e \mbg{\Phi} \mbg{\alpha}_e = \mbr{1}_{n_p} \Rightarrow \mbg{\Phi} \left( \sum_e \mbg{\alpha}_e \right) = \mbr{1}_{n_p}
	\]
	Thus, the question becomes: does $\mbg{\Phi} \mbr{z} = \mbr{1}$ have a solution?  The answer is yes, due to the special properties of the Laplacian matrix, and its eigenfunctions.  Let the eigenvalues of the Laplacian be denoted $\{\mu_i\}$.  Then a result from spectral graph theory \cite{chung1997spectral} shows that if $\mbg{\beta}$ is given by
	\[
	\beta_i =
	\begin{cases}
	\|\mbg{\phi}_i\|_1 & \text{if } \mu_i = 0 \\
	0 & \text{if } \mu_i > 0
	\end{cases}
	\]
	then $\mbg{\Phi}\mbg{\beta} = \mbr{1}$, and it is the unique solution.  This means that the equality constraints can be converted to
	\[
	\sum_e \mbg{\alpha}_e = \mbg{\beta} \quad \Rightarrow \quad \left(\mbr{1}_{n_e} \otimes \mbr{I}_{n_p}\right) \mbg{\alpha} = \mbg{\beta}
	\]
	which quite clearly have a multiplicity of solutions.
	
	Thus, the QP becomes
	\[
	\min_{\mbg{\alpha}} \mbg{\alpha}^T \mbg{\Psi}_{\alpha\alpha} \mbg{\alpha} + \mbg{\theta}_\alpha(\mbr{T})^T \mbg{\alpha}
	\]
	subject to
	\[
	\left(\mbr{1}_{n_e} \otimes \mbr{I}_{n_p}\right) \mbg{\alpha} = \mbg{\beta} \quad \quad \quad \hat{\mbg{\Phi}} \mbg{\alpha} \ge 0
	\]
	At first, this looks very reasonable: the number of variables if $n_b n_e$, which is quite manageable; and the equality constraints have been taken care of.  The problem is the inequality constraints, specifically, the number of such constraints.  Recall that these constraints derive from the non-negativity constraint on each weight, i.e. $w_{ep} \ge 0$ ; thus, there are still $n_e n_p$ such constraints.  Since the complexity of quadratic programming depends on both the number of variables \emph{and} the number of constraints, we will still have a slow algorithm.
	
	\para{Squared Weights}
	As a result, we must reduce the number of inequality constraints.  We ask the following question: what would happen if we dropped these non-negativity constraints?  Examining the various terms of the energy, we see that neither the semantic data term $\mr{E}_1$, nor the spatial smoothness term $\mr{E}_3$, nor the sparsity term $\mr{E}_4$ would have any particular incentive to choose negative weights.  Furthermore, the weight-vote agreement term $\mr{E}_5$ would suffer significantly from choosing negative weights; and the non-collision term $\mr{E}_6$ has no dependence on the weights.  Thus, the only problematic term is the geometric data term $\mr{E}_2 = \mbr{d}(\mbr{T})^T \mbr{w}$.  Quite clearly, this term would be minimized by choosing $w$ as negative as possible.
	
	To deal with this problem, we adopt the following simple workaround.  We change the geometric data term $\mr{E}_2$, to be the following:
	\[
	\mr{E}_2' = \sum_{p, e} w_{ep}^2 D(x_p, T_e \mathcal{X}_e) = \mbr{w}^T \diag(\mbr{d}(\mbr{T})) \mbr{w}
	\]
	That is, we replace the weights in the weighted ICP with the \emph{squared weights}.  Now, choosing large negative weights will pose a disadvantage.  In fact, even small negative terms would tend to raise $\mr{E}_2'$, when coupled with the constraint $\sum_e w_{ep} = 1$; this is because if one term is negative, that means that the others must sum to more than 1.
	
	Making this change yields the following changes in the problem parameters:
	%\footnote{At this stage, we suppress explicit $\mbr{T}$ dependence of the parameters for greater clarity, as $\mbr{T}$ is fixed at this stage, and other stages of the algorithm are not affected.}
	\[
	\mbg{\Psi}_{ww}'(\mbr{T}) = \tilde{\mbg{\Psi}}_{ww}' + \lambda_2 \diag(\mbr{d}(\mbr{T})) \quad \quad \mbg{\theta}_w'(\mbr{T}) = \mbg{\theta}_w(\mbr{T}) - \lambda_2 \mbr{d}(\mbr{T})
	\]
	and the inequality constraints are removed.  That is, taking $\mbg{\Psi}_{\alpha\alpha}'(\mbr{T}) = \hat{\mbg{\Phi}}^T \tilde{\mbg{\Psi}}_{ww}'(\mbr{T}) \hat{\mbg{\Phi}}$ and $\mbg{\theta}_\alpha'(\mbr{T}) = \mbg{\theta}_w'(\mbr{T}) - \mbg{\Psi}_{vv}(\mbr{T}) \mbr{v}$ we solve
	\begin{equation}
		\min_{\mbg{\alpha}} \mbg{\alpha}^T \mbg{\Psi}_{\alpha\alpha}'(\mbr{T}) \mbg{\alpha} + \mbg{\theta}_\alpha'(\mbr{T})^T \mbg{\alpha} \quad \text{s.t.} \quad \left(\mbr{1}_{n_e} \otimes \mbr{I}_{n_p}\right) \mbg{\alpha} = \mbg{\beta}
		\label{eq:min_alpha}
	\end{equation}
	In this case, the QP can actually be solved as a linear system, as it is the result of minimizing a positive semi-definite quadratic form subject to a linear system.  This yields a very fast algorithm in practice.
	
	\para{The Accelerated Algorithm}  The faster algorithm is identical to Algorithm \ref{alg:main}, with one change: the segmentation step now involves solving (\ref{eq:min_alpha}) for $\mbg{\alpha}$ instead of (\ref{eq:min_w}) for $\mbr{w}$.
	
	\section{Results}
	\label{sec:Results}
	To evaluate the performance of ASIST, we conducted several experiments using a variety of datasets including both our own synthetic and scanned scenes as well as scenes acquired by Nan \etal \cite{Nan2012} and Li \etal \cite{Li2015}. Overall our datasets contain scans acquired by four different types of sensors: Kinect v1, Kinect v2, Mantis Vision, and Google Tango.
	
	This section is organized as follows. In Section \ref{Experimental_Settings} we detail the parameters of ASIST, such as the number of iterations, weights for the different energy terms, etc. In Section \ref{Synthetic_Dataset} we evaluate ASIST on a dataset of scenes containing synthetic models obtained from ModelNet \cite{Song2015}. We evaluate our performance both quantitatively and qualitatively, and discuss success and failure cases. In \ref{Fused_Scans} we introduce a small scanned dataset containing scenes we acquired using Kinect v2 and Tango, and discuss the performance of ASIST on these scenes. Sections \ref{Li_Results} and \ref{Nan_Results} include comparison with two different algorithms from the recent literature \cite{Li2015,Nan2012}. Since there was no available implementation for either method, we ran ASIST on scanned datasets supplied by the authors on which they reported their performance. In both cases we found ASIST to give comparable results. 
	
	\subsection{Experimental Settings}
	\label{Experimental_Settings}
	\para{General Parameters} In all experiments we ran our algorithms for $N_{out}=5$, $N_{in}=2$, and $N_{IRLS}=5$ iterations.  The $\|\cdot\|_\ell$ sparsity inducing norm was chosen as $\ell = 0.1$.  The energy term coefficients were taken to be $\lambda_3=100$, $\lambda_4=10$, and $\lambda_5=1$. As was described in Section \ref{sec:Implementation}, $\lambda_6$ was increased at each outer iteration, taking on values of $1, 5, 10, 10^2, 10^3, 10^9$. The rest of the parameters varied by experiment, and were set as shown in Table \ref{table_params}.
	
	\para{Forest Training} In all experiments we trained a forest with $9$ trees of depth $10$. At each node a pool of a $1,000$ random split functions were generated and the one maximizing the Shannon information gain was selected. Training was stopped if the information gain was below $0.05$ or if less than $30$ samples reached the node. Each tree was trained of a random set of $6\times 10^5$ cells selected out of models from ModelNet \cite{Song2015}.
	
	The average execution time per scene using unoptimized MATLAB code on an Intel Xeon $E5620$ $2.4GHz$ is about $10$ minutes.
	
	\begin{table}[h]
		\begin{tabular}{| l | c | c | c | c |}
			\hline
			&    $\lambda_1$      	&    $\lambda_2$    & $D_{clutter}$ &  $Voxel [cm]$ \\
			\hline
			Synethetic scenes 					& $1$ 					& $1$ 				& $10$ 				 & $7.5$ \\
			\hline
			Data of Nan \etal \cite{Nan2012} 	& $10$ 					& $1$ 				& $20$				 & $2.5$ \\		
			\hline
			Data of Li \etal \cite{Li2015} 		& $10$ 					& $10$ 				& $10$ 				 & $2.5$ \\
			\hline
		\end{tabular}
		\caption{\small \textbf{Parameter settings}. Reported are only the parameters that vary across experiments. See discussion in the text.}
		\label{table_params}
	\end{table}

	\subsection{Synthetic Dataset}
	\label{Synthetic_Dataset}
	\para{Description of the Dataset}
	As an initial experiment, we evaluated the performance of ASIST on a synthetic benchmark.  The benchmark comprised $30$ scenes each containing a random collection of objects taken from the test portion of the ModelNet \cite{Song2015} dataset (recall that ASIST was trained on the non-overlapping training set of ModelNet). The objects were positioned at random non-overlapping locations within the scene.  Objects from the following five classes were used in our benchmark: chair, table, toilet, sofa, and bed.  Each scene contained either one or two objects from each class. The scene thus consists of a mesh, and is annotated additionally with a set of ground truth bounding boxes and their class labels. Note that the bounding boxes are not axis-aligned; rather, they are aligned tightly to the objects they surround.  
	
	\para{Evaluation criteria}
	In all experiments we computed precision and recall using the bounding boxes; an overlap is considered to have occurred if the Intersection over Union (IoU) score, defined as the ratio between the intersection and union volumes of the bounding boxes \cite{everingham2012pascal}, is greater than $0.25$.  We define two measures of relevance and, consequently, two types of precision and recall: \textit{semantic}, for which the replaced and the replacing objects are deemed relevant if they have matching class labels, and \textit{geometric}, which simply looks for overlap disregarding the class labels of the objects. Semantic precision and recall is reported on a per-class basis, whereas geometric precision and recall is given as an aggregate for the entire dataset.
	
	We intend to release our synthetic benchmark to the research community.
	
	\para{Results}
	\label{Synthetic_Results}
	A quantitative evaluation of the performance of ASIST on the synthetic benchmark is given in Table \ref{table_synthetic_PR}. For each of the five classes, the semantic precision, recall, and $F_1$ scores are reported. Note that the semantic precisions vary between $0.91$ and $1$, with a mean precision of $0.97$ across all classes; while the semantic recalls vary between $0.94$ and $0.98$, with a mean recall of $0.96$ across all classes.  The corresponding $F_1 = 2\cdot\frac{precision \cdot recall}{precision + recall}$ scores range between $0.94$ and $0.99$, with a mean value of $0.96$.  In addition, the geometric precision and recall (which do not account for class labels) are $1$ and $0.99$, with the geometric $F_1$ score of $0.99$.
	
\begin{figure*}
		\label{table_synthetic_results}
		\centering
		\begin{tabular}{ c c c }
			\includegraphics[width = 0.3\textwidth]{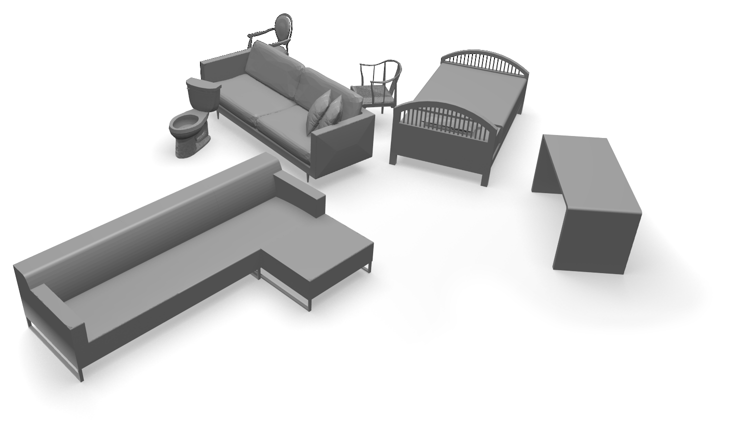}  &           
			\includegraphics[width = 0.3\textwidth]{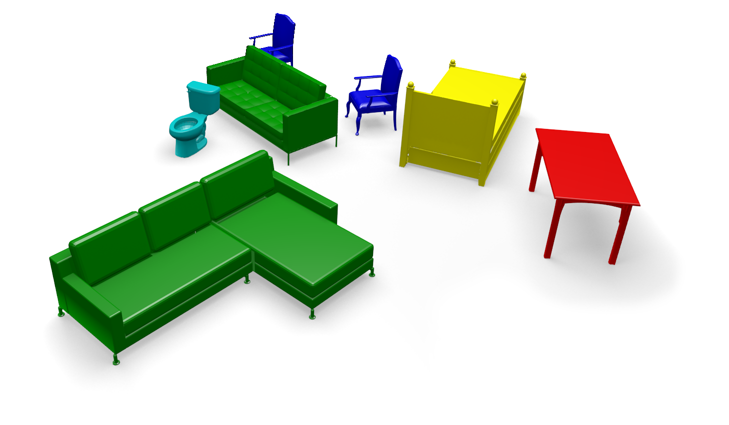}  &
			\includegraphics[width = 0.3\textwidth]{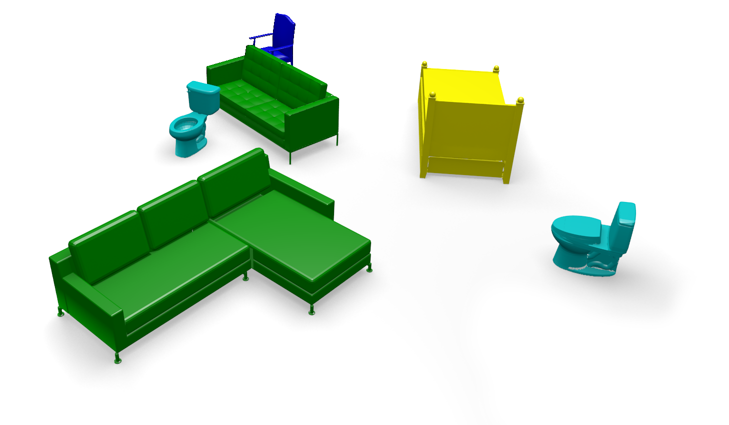}\\

			\includegraphics[width = 0.3\textwidth]{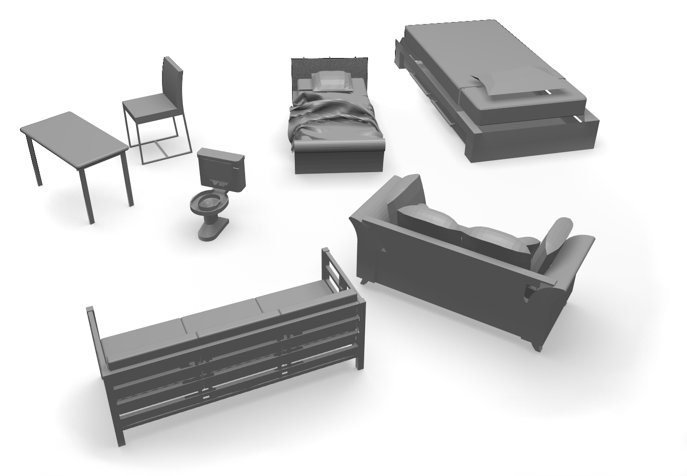}  &
			\includegraphics[width = 0.3\textwidth]{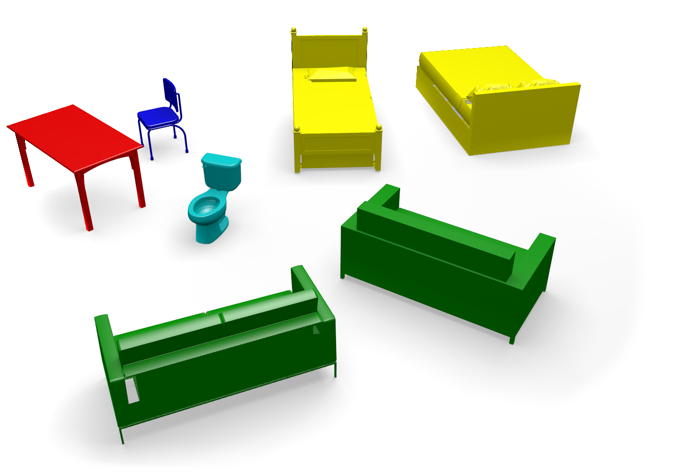}  &
			\includegraphics[width = 0.3\textwidth]{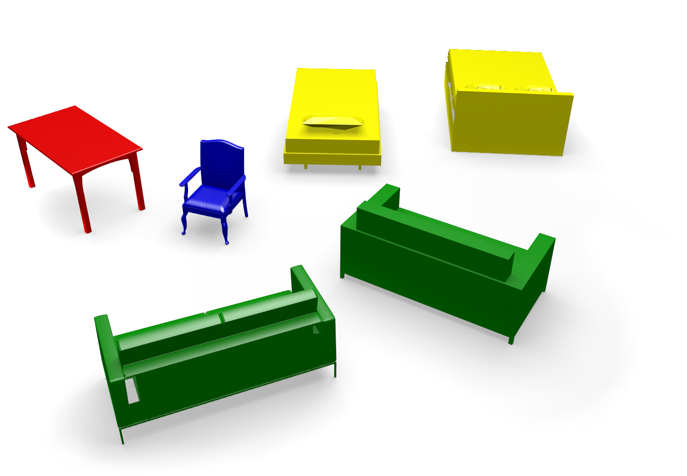}\\

			\includegraphics[width = 0.3\textwidth]{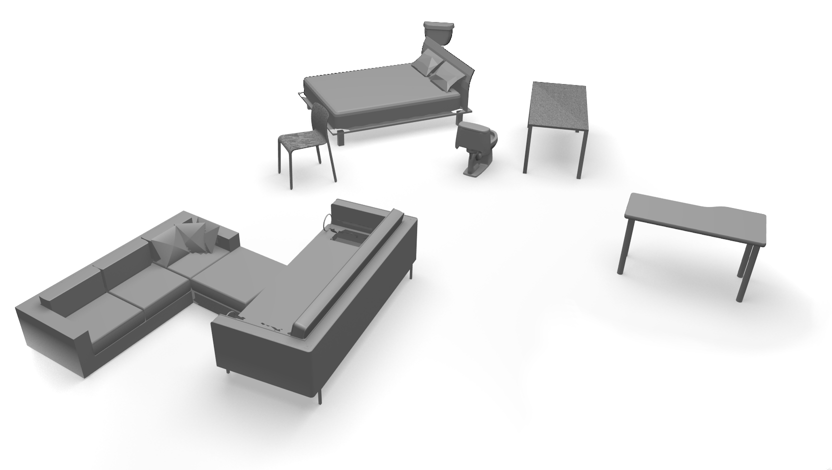}  &           
			\includegraphics[width = 0.3\textwidth]{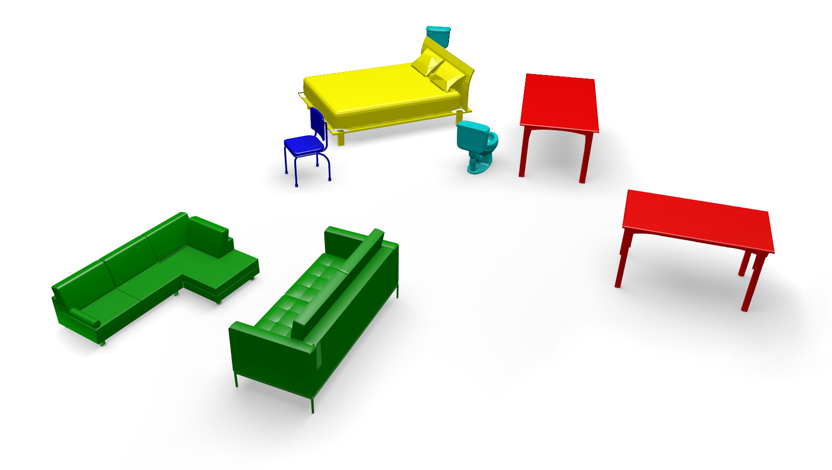}  &
			\includegraphics[width = 0.3\textwidth]{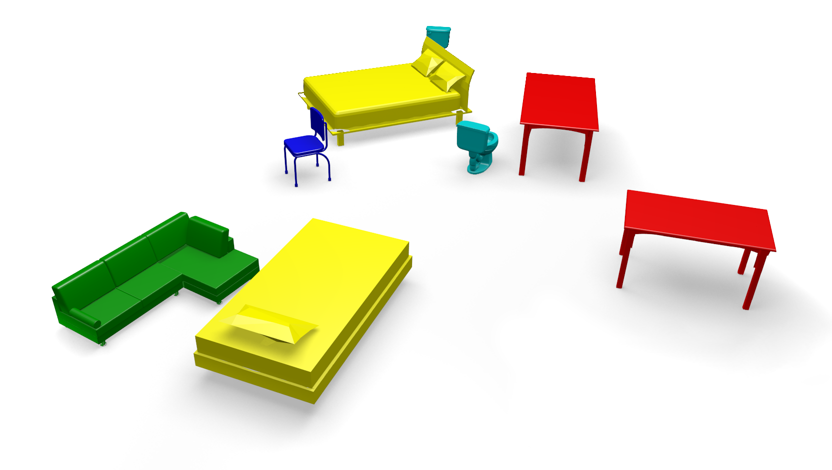}\\

			(a) & (b) & (c) \\
		\end{tabular}   \\
		\caption{\textbf{Results on synthetic scenes.} Column (a) shows the original scene; (b) shows the output of ASIST for $N_{out} = 5$; (c) shows the output of ASIST for $N_{out} = 1$. Each row shows the synthetic scene in gray; and the output of the ASIST algorithm, in which the objects to be inserted are rendered in color-coded scheme according to object class (blue for chairs, red for tables, cyan for toilets, green for sofas, and yellow for beds). The two colored columns (b) and (c) represent the two ASIST configurations with $N_{out}=5$ and $N_{out}=1$ respectively.}		
	\end{figure*}

	In our experiments, ASIST is executed for $N_{out}=5$ outer loop iterations.  We compare our results with those for which ASIST is run for $N_{out}=1$. In this setting we assign a very high value for $\lambda_6$.  Given that the number of inner iterations is rather small ($N_{in} = 2$), $N_{out}=1$ entails relatively little interaction between the segmentation, registration, and voting steps. Thus, comparing these two different settings of ASIST gives an indication of the importance of the unified approach to semantic segmentation and object replacement on which the ASIST algorithm is based.
	Several example scenes are shown in Figure \ref{table_synthetic_results}. As can be observed from these examples, ASIST succeeds in properly identifying both the class and pose of the scene objects, and in finding a geometrically close substitute.
	
	It is also evident from Figure \ref{table_synthetic_results} that the number of iteration affecs the performance of ASIST. It can be seen in first example presented in Figure \ref{table_synthetic_results} that while ASIST with $N_{out}=5$ returns a perfect result, ASIST with $N_{out}=1$ replaces a table with a toilet (bottom right part of the scene), completely misses a chair (center of the scene), and places the bed at a wrong orientation (top right corner of the scene). In the second example ASIST with $N_{out}=1$ replaces a toilet with a chair (center of the scene), completely misses a chair (top left corner of the scene), and once again positions the bed at a wrong orientation (top right corner of the scene). In the third example it replaces a sofa with a bed (bottom part of the scene).
	
\begin{figure*}
		\label{table_synthetic_failures}
		\centering
		\begin{tabular}{ c c c }
			\includegraphics[width = 0.3\textwidth]{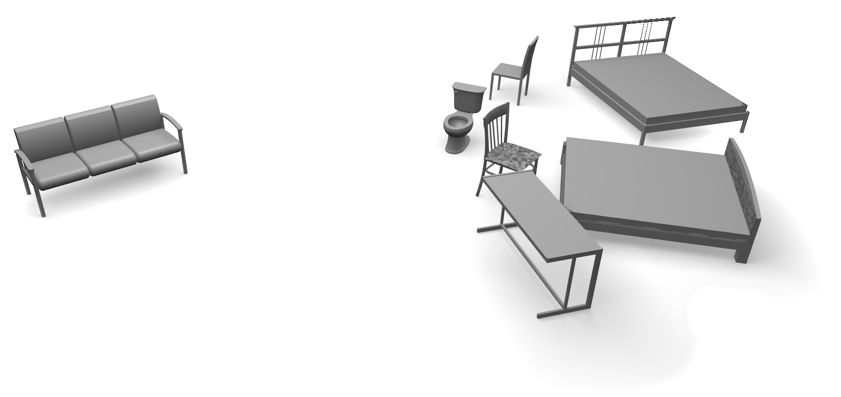}  &           
			\includegraphics[width = 0.3\textwidth]{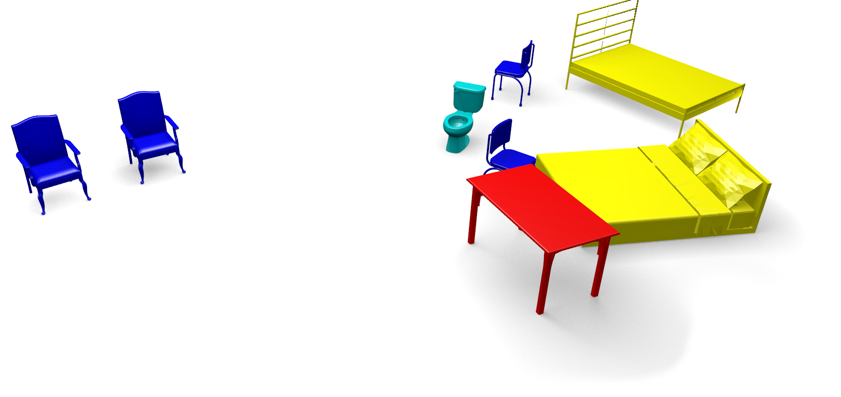}  &
			\includegraphics[width = 0.3\textwidth]{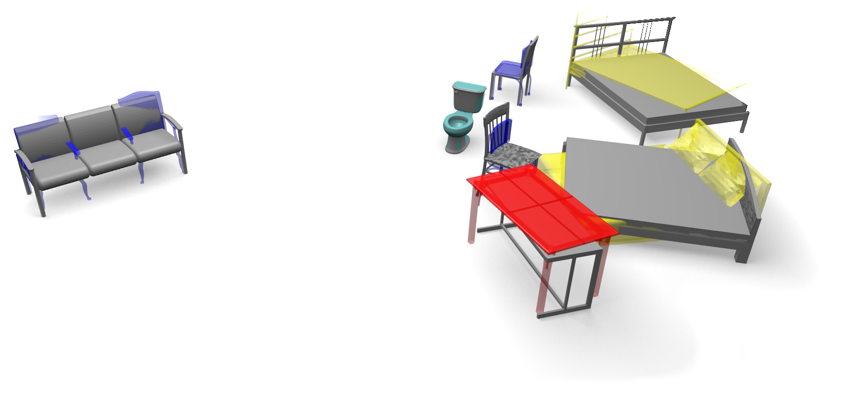} \\
			
			\includegraphics[width = 0.3\textwidth]{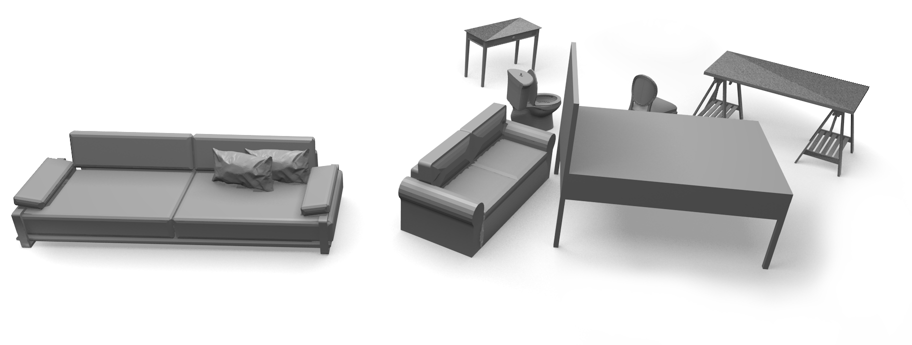}  &
			\includegraphics[width = 0.3\textwidth]{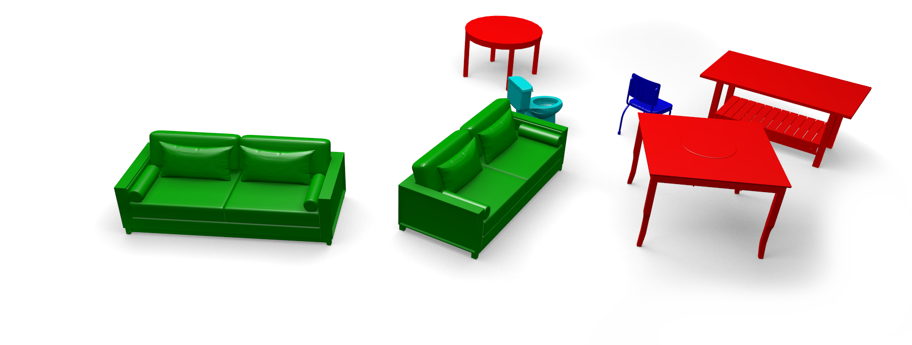}  &
			\includegraphics[width = 0.3\textwidth]{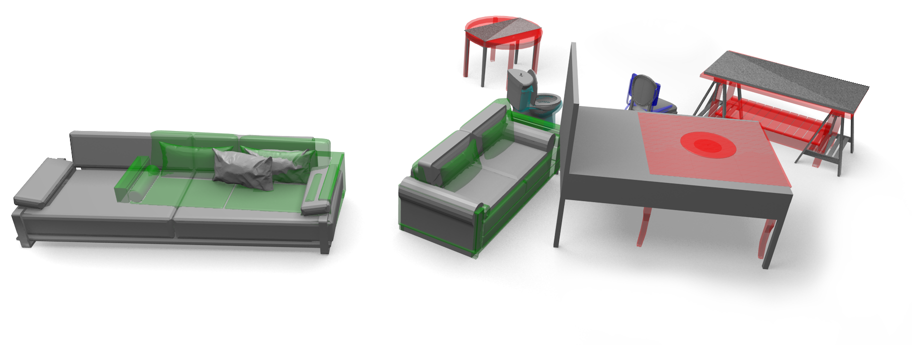} \\
			
			(a) & (b) & (c) \\
		\end{tabular}   \\
		\caption{\textbf{Failure cases.} Column (a) shows the original scene; (b) shows the output of our algorithm after five iterations; (c) shows the scene and the algorithm's output superimposed.}		
	\end{figure*}

	Failure cases of the algorithm are shown in Figure \ref{table_synthetic_failures}.  The example in the top row shows two different kinds of errors.  First, a sofa in the scene is replaced by two chairs (top left of the scene).  Second, both beds (in yellow, top right of the scene) are roughly recovered, but their pose is incorrect.  In the example in the bottom row, a bed is replaced with a table (in red, bottom right); the sofa at the bottom left is replaced with a considerably smaller sofa; and a rectangular table (top middle) is replaced with circular table.  These sorts of errors are characteristic of ASIST's failure modes.
	
	The precision, recall, and $F_1$ scores for $N_{out}=1$ are reported in Table \ref{table_synthetic_PR}.  The mean semantic precision across classes for a single iteration is $0.95$; when compared to $0.97$ for five iterations, this demonstrates a modest improvement for the full algorithm.  The improvement is much clearer in examining the recall scores, where the mean semantic recall across for a single iteration is $0.85$, compared to $0.96$ for five iterations.

	Thus, we may conclude that the unified approach of ASIST to semantic segmentation and object replacement is indeed important; as it provides crucial improvements to the algorithm's performance.

	\begin{table}
		\small
		\centering
		\begin{tabular}{| l | c | c | c | c | c | c |}\hline
			\textbf{Measure} \textbackslash \textbf{Class}
			& bed & chair & sofa  & table  & toilet  & geo     \\
			\hline
			Prec. ($N_{out}=5$)     & \tbf{0.98}  & \tbf{0.96}    & \tbf{1}   & \tbf{0.91}    & \tbf{1}       & \tbf{1}       \\
			Prec. ($N_{out}=1$)     & 0.97        & 0.93          & 0.98      & 0.91          & 0.97          & 0.99          \\
			\hline
			Rec ($N_{out}=5$)        & \tbf{0.95}  & \tbf{0.94}    & \tbf{0.96}& \tbf{0.98}    & \tbf{0.98}    & \tbf{0.99}    \\
			Rec ($N_{out}=1$)        & 0.83        & 0.88          & 0.91      & 0.78          & 0.88          & 0.89          \\
			\hline
			$F_1$ ($N_{out}=5$)        & \tbf{0.96}  & \tbf{0.95}    & \tbf{0.98}& \tbf{0.94}    & \tbf{0.99}    & \tbf{0.99}    \\
			$F_1$ ($N_{out}=1$)        & 0.9         & 0.91          & 0.95      & 0.84          & 0.92          & 0.94          \\
			\hline
		\end{tabular}
		\caption{\textbf{Synthetic Benchmark Results}. Semantic and geometric precision, recall and $F_1$ scores for each of the $5$ classes over the $30$ benchmark scenes.}
		\label{table_synthetic_PR}
	\end{table}

	\subsection{Fused Scans}
	\label{Fused_Scans}
\begin{figure*}
		\label{fusion_scenes}
		\centering
		\setlength{\tabcolsep}{0 pt}
		\begin{tabular}{c c c}
			\includegraphics[width = 0.3\textwidth]{lior_dinning_scan_crop.png} &
			\includegraphics[width = 0.3\textwidth]{lior_dinning_models_crop.png} &
			\includegraphics[width = 0.3\textwidth]{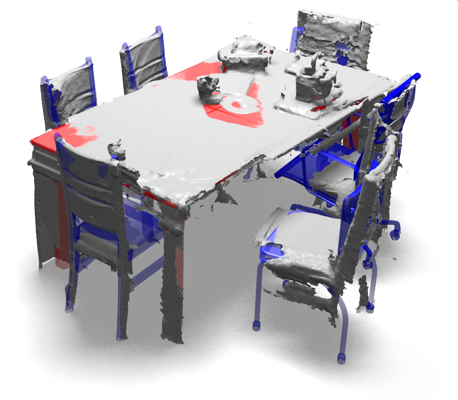} \\
			
			\includegraphics[width = 0.3\textwidth]{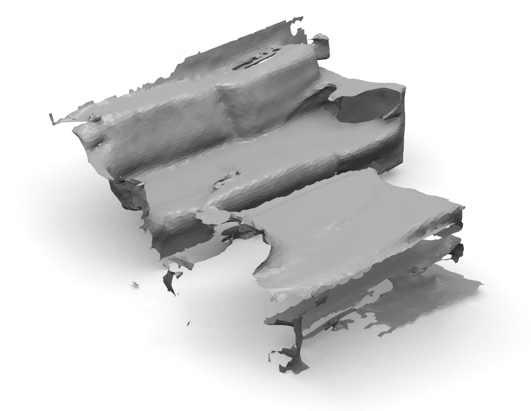} &		
			\includegraphics[width = 0.3\textwidth]{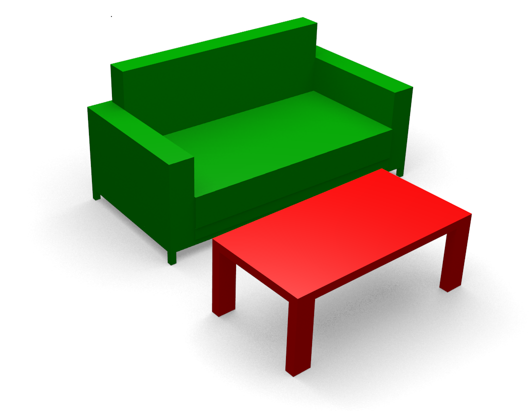} &
			\includegraphics[width = 0.3\textwidth]{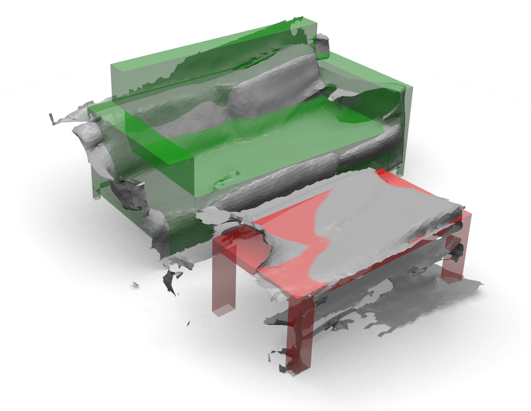} \\
			
			\includegraphics[width = 0.3\textwidth]{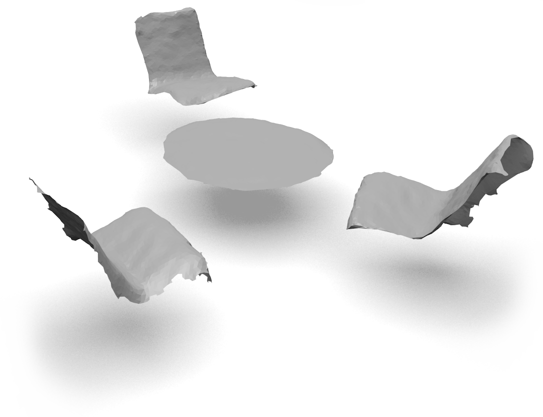} &		
			\includegraphics[width = 0.3\textwidth]{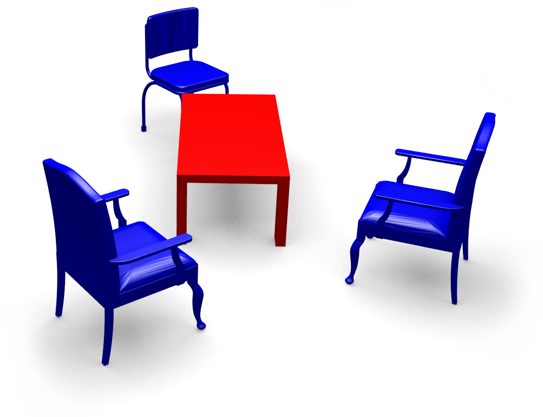} &
			\includegraphics[width = 0.3\textwidth]{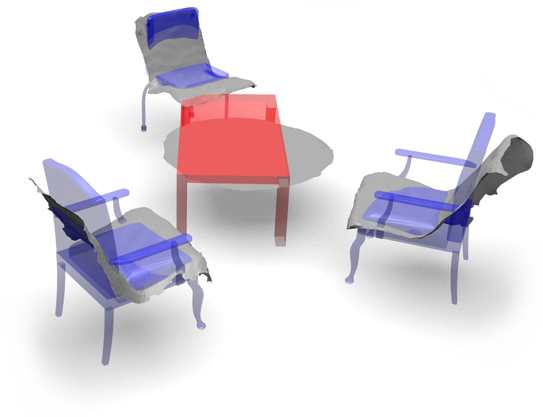} \\
			
			\includegraphics[width = 0.3\textwidth]{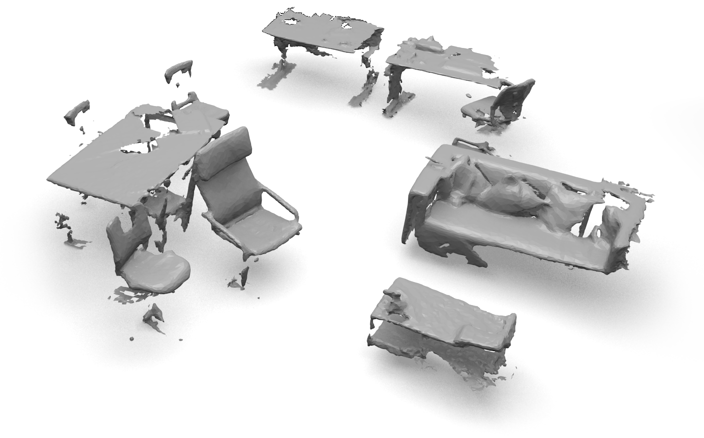} &
			\includegraphics[width = 0.3\textwidth]{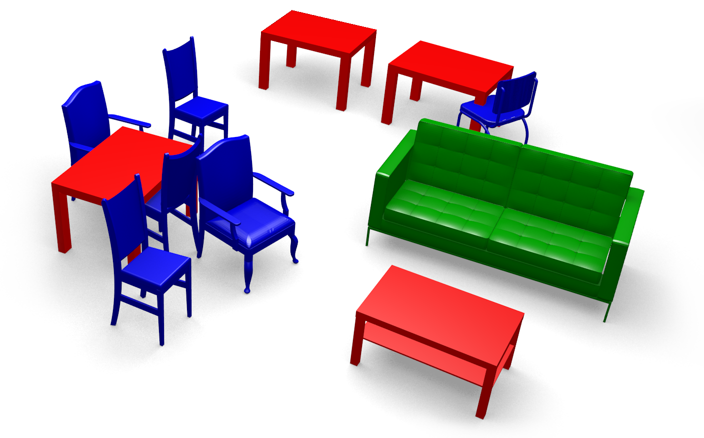} &
			\includegraphics[width = 0.3\textwidth]{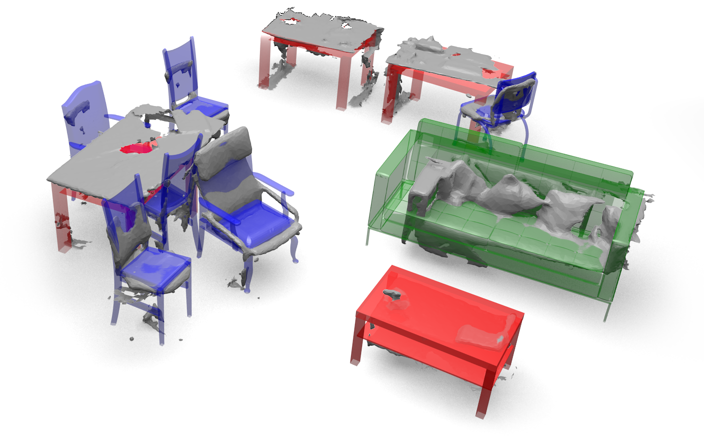} \\
			
			(a) & (b) & (c) \\
			
		\end{tabular}
		\caption{\textbf{ASIST using a Kinect sensor.} Column (a) scanned scene; (b) ASISTs' result; (c) overlay of the previous two. }		
	\end{figure*}

\begin{figure*}
		\label{tango_v3}
		\centering
		\begin{tabular}{c c c}
			\includegraphics[width = 0.3\textwidth]{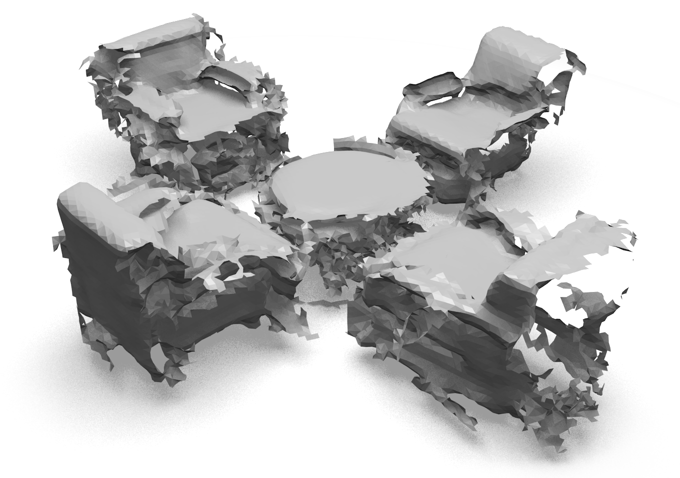} &		
			\includegraphics[width = 0.3\textwidth]{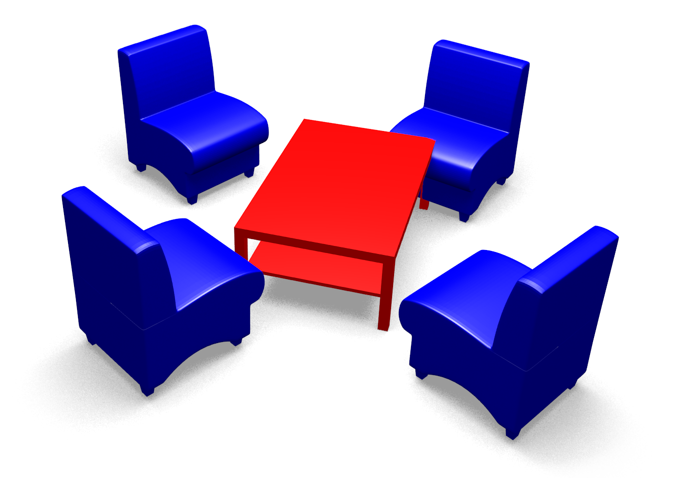} &
			\includegraphics[width = 0.3\textwidth]{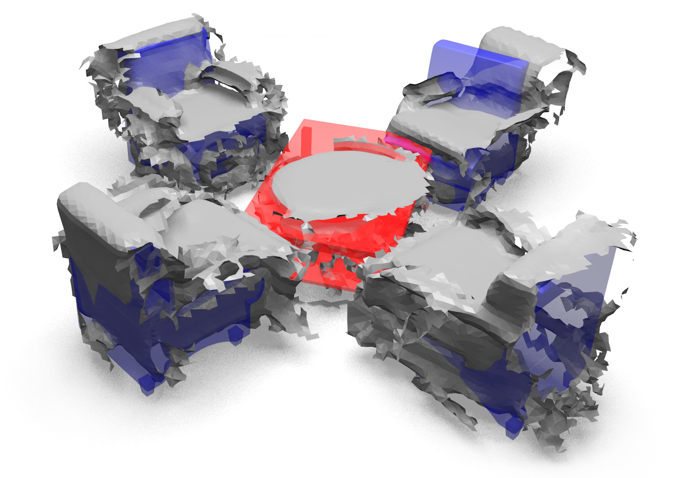} \\
			
			(a) & (b) & (c) \\
		\end{tabular}
		\caption{\textbf{ASIST using a Google Tango sensor.} Column (a) scanned scene; (b) ASISTs' result; (c) overlay of the previous two.}		
	\end{figure*}

\begin{figure*}
		\label{tango_v2}
		\centering
		\begin{tabular}{c c c}
			\includegraphics[width = 0.3\textwidth]{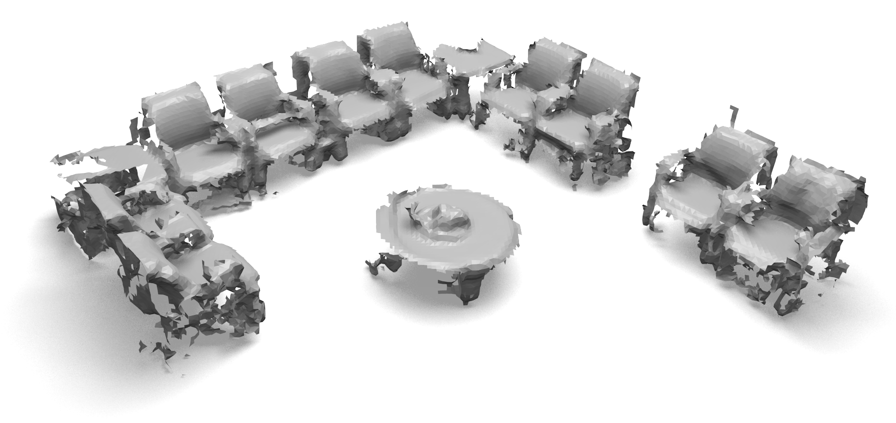} &		
			\includegraphics[width = 0.3\textwidth]{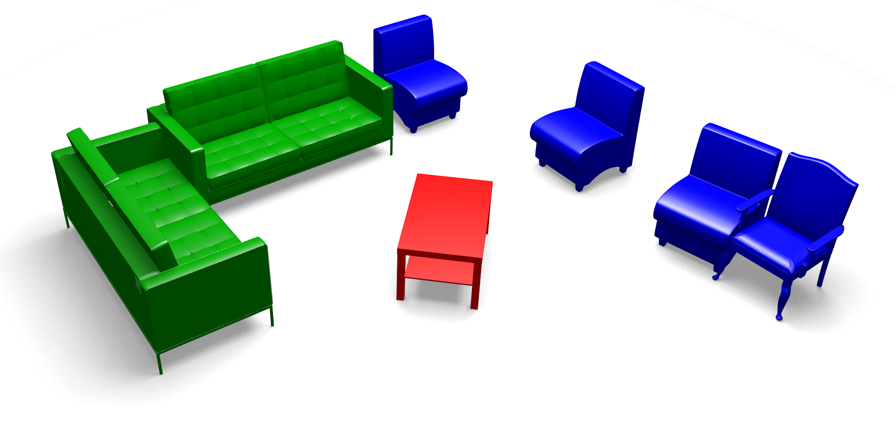} &
			\includegraphics[width = 0.3\textwidth]{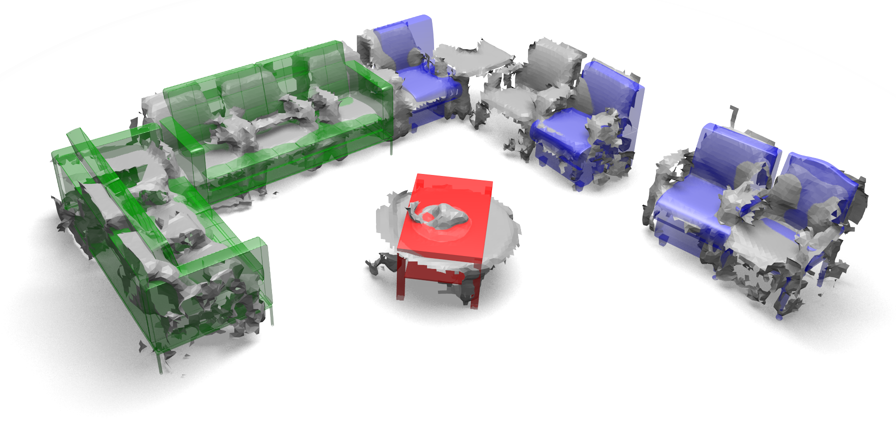} \\
			
			(a) & (b) & (c) \\
		\end{tabular}	
		\caption{\textbf{ASISTs' failure on a scene scanned with Google Tango.} (a) scanned scene; (b) ASIST result; (c) overlay of the previous two.}		
	\end{figure*}

	\para{Description of the Dataset}
	As explained in previous sections, the ASIST algorithm is designed to operate on point clouds obtained by a fusion of scans from multiple directions. To evaluate its performance in these scenarios, we collected several such fused scans using two types of sensors: Kinect v2 based on time-of-flight technology, and Tango based on triangulation \cite{projectTango}. We built a small collection of six indoor scenes, four of which were obtained using Kinect Fusion \cite{izadi2011kinectfusion} and two using Tango. The scenes are presented in the leftmost columns of Figures \ref{fusion_scenes}-\ref{tango_v2} where it can be seen that even though the acquisition was done from multiple viewpoints, it consists of many holes, mainly due to occlusions. Data acquired using Tango is also much noisier, with approximately half the average point density of the scenes acquired with Kinect v2. The dataset consists a total of $29$ chairs, $9$ tables and $2$ sofas. 
	
\begin{figure*}
		\label{pipe_before_after_spectral}
		\centering
		\begin{tabular}{ c  c  c}
			\includegraphics[width = 0.3\textwidth]{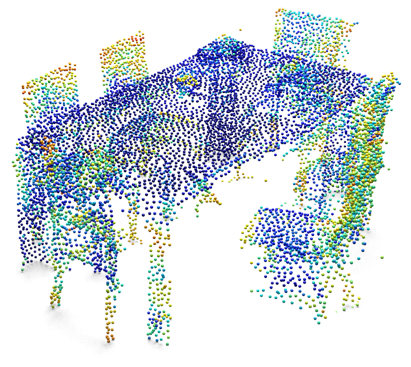}  &
			\includegraphics[width = 0.3\textwidth]{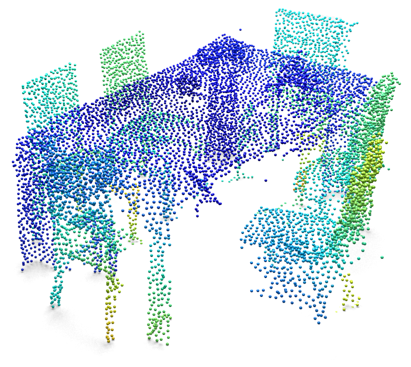}  & 
			\includegraphics[width = 0.3\textwidth]{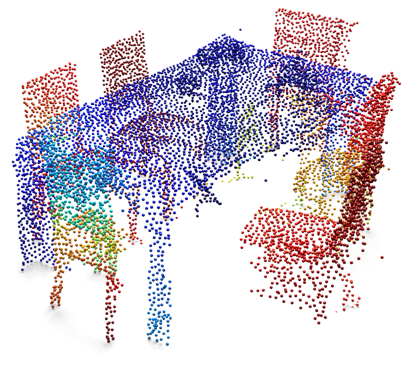} \\	
			\includegraphics[width = 0.3\textwidth]{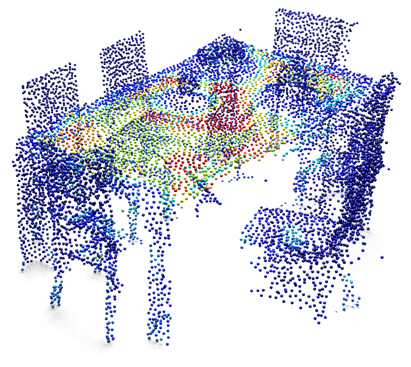}  &		
			\includegraphics[width = 0.3\textwidth]{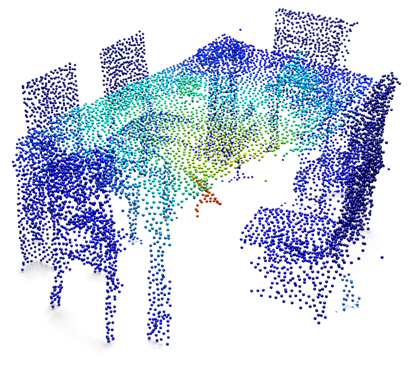}  & 
			\includegraphics[width = 0.3\textwidth]{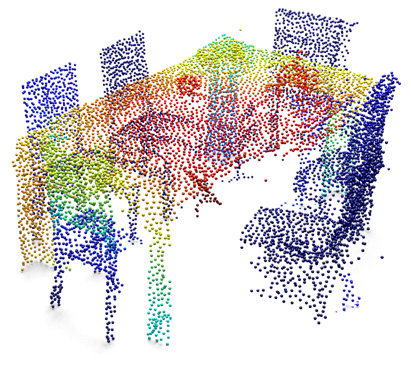} \\
			(a) & (b) & (c) \\
		\end{tabular}
		\caption{\textbf{Pipeline illustration, Part 1.} Per point semantic class segmentation at different steps of ASIST, for the chair class (top row) and table class (bottom row). (a) Initial (forest) labeling; (b) after spectral representation; (c) final segmentation.}		
	\end{figure*}

\begin{figure*}
		\label{pipe_exm_locations}
		\centering
		\begin{tabular}{ c  c  c  c }
			\multicolumn{4}{c}{\includegraphics[width = 0.6\textwidth]{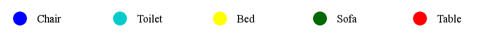}}  \\
			\includegraphics[width = 0.23\textwidth]{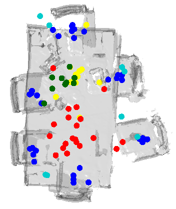} &
			\includegraphics[width = 0.23\textwidth]{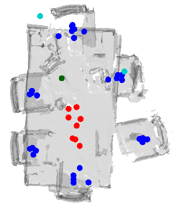} &
			\includegraphics[width = 0.23\textwidth]{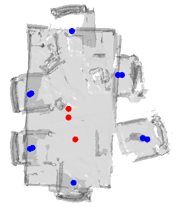} &		
			\includegraphics[width = 0.23\textwidth]{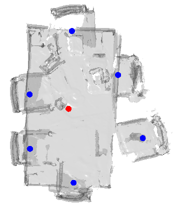} \\		
			
			Initialization & Iteration 1 & Iteration 3 & Iteration 5 \\
		\end{tabular}
		\caption{\textbf{Pipeline illustration, Part 2.} Exemplar locations: (a) at initialization; (b) after a single iteration; (c) after $3$ iterations; and (d) after $5$ iterations. }		
	\end{figure*}

	\para{Pipeline Illustration}
	In order to get a better understanding of the different steps of ASIST, we present intermediate results on the ``dining table'' scene shown in the top row of Figure \ref{fusion_scenes}. Figures \ref{pipe_before_after_spectral} and \ref{pipe_exm_locations} show intermediate outputs of the algorithm pipeline. The semantic segmentation is presented in Figure \ref{pipe_before_after_spectral}. Each point in the scene's point cloud is assigned a semantic class confidence, which is just the sum of the weights of all exemplars belonging to the same class. The points are colored according to a heat-map, where red means high confidence and blue signifies low confidence. The leftmost column depicts the labeling achieved by the random forest at initialization for the chair class (top row) and for the table class (bottom row). In this example we used the six-class forest (clutter, chair, toilet, bed, sofa and table) but since the scene contains only chairs and a table, for the sake of clarity we present only the relevant weights. It can be seen that the classifier succeeds in locating unique characteristics of the class such as the height of the plane in the table case and legs and backrest of the chairs. However, this results is far from being accurate enough for performing the scene transformation.
	
	Figure \ref{pipe_before_after_spectral} column (b) shows the result of using the spectral basis representation. Other than having the obvious benefit of reducing the number of optimization variables significantly, this low-pass filtering type of action has the effect of smoothing out the per-point labeling, thereby assisting the Mean Shift algorithm to avoid getting stuck in local modes. Finally, column (c) shows the final labeling achieved by ASIST. It can be seen than the objects points assume a much higher value of their corresponding class confidence. This demonstrates the power of the alternations done by the algorithm, i.e. segmentation contributes to better registration and vice versa. While this intermediate result is clearly not prefect, it is sufficiently accurate in order to perform a perfect replacement.
	
	The evolution of exemplars though the course of $N_{out} = 5$ iterations is shown in Figure \ref{pipe_exm_locations}. The scene is shown in transparent gray from a top viewpoint and each exemplar is represented as a full circle located at its bounding box's center. Each circle is assigned a color according to its class membership, using the same color code as described in \ref{Synthetic_Results}. Column (a) depicts all the exemplar locations at initialization, i.e. after each exemplar has been placed at its corresponding class' Mean Shift mode, and was then registered to the scene using several weighted ICP's, each with a different initial rotation around $z$-axis. The location shown is of the exemplar position as output from the ICP that resulted in the smallest one-sided Hausdorff distance between the exemplar and the scene. Columns (b)-(d) show the locations of the exemplars $e$ for which the vote $v_e$ is positive and the aggregation of weights exceed a small threshold (this is the same criterion denoted by $\epsilon_{rep}$ in Algorithm \ref{alg:main}).
	
	Figure \ref{pipe_exm_locations} demonstrates nicely how the large number of initial candidates is reduced at each step until we are left with only a final subset of exemplars which are both non-overlapping and have high confidence values. Increasing the value of $\lambda_6$ in a moderate fashion results in gradually removing the exemplars with low confidence while keeping the more ambiguous ones even if they are overlapping, thus leaving the hard selection for later steps where the best fitting exemplar presents a more significant advantage as compared to the others. Observe how after a single iteration the candidate exemplars belonging to the bed class have already been excluded, and after the third iteration only exemplars from the correct classes remain for ASIST to choose from. This gives good intuition for why ASIST tends to converge in practice to the correct solution even though it is not a convex problem; it is because the voting step (see Algorithm \ref{alg:main}) is initialized close to the global solution and its non-convex term, $\lambda_6 \mr{E}_6$ (the non-collision term), increases gradually.
	
	Finally, the transformed scene is shown in the top row of Figure \ref{fusion_scenes}.
	
	%
	% Exemplar locations
	%
	% Final segmentation
	%\begin{figure*}
	%	\centering
	%	\begin{tabular}{ c  c }
	%		\includegraphics[width = 0.5\textwidth]{pipeline_dining_chairs_final_crop.png} &
	%		\includegraphics[width = 0.5\textwidth]{pipeline_dining_table_final_crop.png} \\
	%	\end{tabular}
	%	\label{pipe_final_seg}
	%	\caption{Final class segmentation for classes: (a) Chair and (b) Table }
	%\end{figure*}
	
	\para{Results}
	In this experiment we present the result of ASIST on six scanned scenes, with floors removed.  As can be seen in Figures \ref{fusion_scenes}-\ref{tango_v2} we get a perfect result in terms of precision and recall on five out of the six scenes. We now elaborate on these results as well as the interesting failure case shown in Figure \ref{tango_v2}. 
	
	As can be seen in Figure \ref{fusion_scenes}, the scans obtained using the Kinect sensor are of a reasonable resolution. Nevertheless they contain a non negligible amount of clutter (see the top of the dining table for example) and noise (see the short coffee table in the second row). These are handled though our clutter-class and the exemplars' geometry. Recall that due to the alternating fashion of the algorithm the registration, while relying on the segmentation, also aids in improving it.
	
	Taking a closer look at the first example shown in Figure \ref{fusion_scenes}, the clutter on the table can clearly be seen, while additionally some of the chairs are missing their legs. ASIST handles these issues and returns the correct result as a result of its unified approach. The second example shows how ASIST successfully deals with large occlusions: a large part of the sofa and table are missing from the scan. The third example also contains serious occlusions.  It is evident in this case that ASIST returns a rectangular table rather than a round one and chairs that are less reclined than the ones in the scene.  This is due to the lack of a better exemplar in the database; nevertheless the result is faithful to the semantics of the scene. The fourth example emphasizes how ASIST handles large and crowded scenes with many different objects and classes, where again, some of the scanned objects have missing parts such as legs and seats. Nevertheless, all objects are recovered correctly with minor geometric errors such as the length of the couch and the leftmost table.
	
	Figure \ref{tango_v3} shows an example of a scene scanned with the Google Tango sensor. The scan resolution is much lower compared to the Kinect scenes with about half of the spatial resolution. Large parts of the scan are missing and those present are extremely noisy. Yet, these acquisition imperfections are still robustly handled by our algorithm.
	
	In Figure \ref{tango_v2} we present a failure case of ASIST on a scene scanned using Google Tango. The errors can be divided into two types: geometric, such as the returned table being rectangular instead of round and a few missing chairs at the top right corner of the scene; and semantic, such as the chairs on the left side of the scene that were replaced by sofas. Although this is a semantic error the two class types are relatively close semantically and this sort of error could probably be acceptable in some VR applications.

	%\begin{enumerate}
	%	\item Table of precision/recall (geometric and semantic) for each of the object categories. \orl{Let's discuss this.}
	%	\item In the same table, the same quantities computed for the one-shot (single iteration) algorithm.
	%	\item Images of several (6-10) sections of successful scenes. \orl{DONE}
	%	\item Images of the same (6-10) sections of scenes, done by the one-shot algorithm. 
	%	\item Images of 2-3 sections of unsuccessful scenes, in which there is an interesting failure case. 
	%	\item Figure showing the weights and v during iterations (init (before/after spectral decomposition), first and final)
	%	
	%	
	%\end{enumerate}
	
	\subsection{Comparison to Li \etal\cite{Li2015}}
	\label{Li_Results}
\begin{figure*}
		\label{ikea_scenes}
		\centering
		\begin{tabular}{ c c c c}
			
			\multicolumn{4}{c}{ \includegraphics[width = 0.9\textwidth]{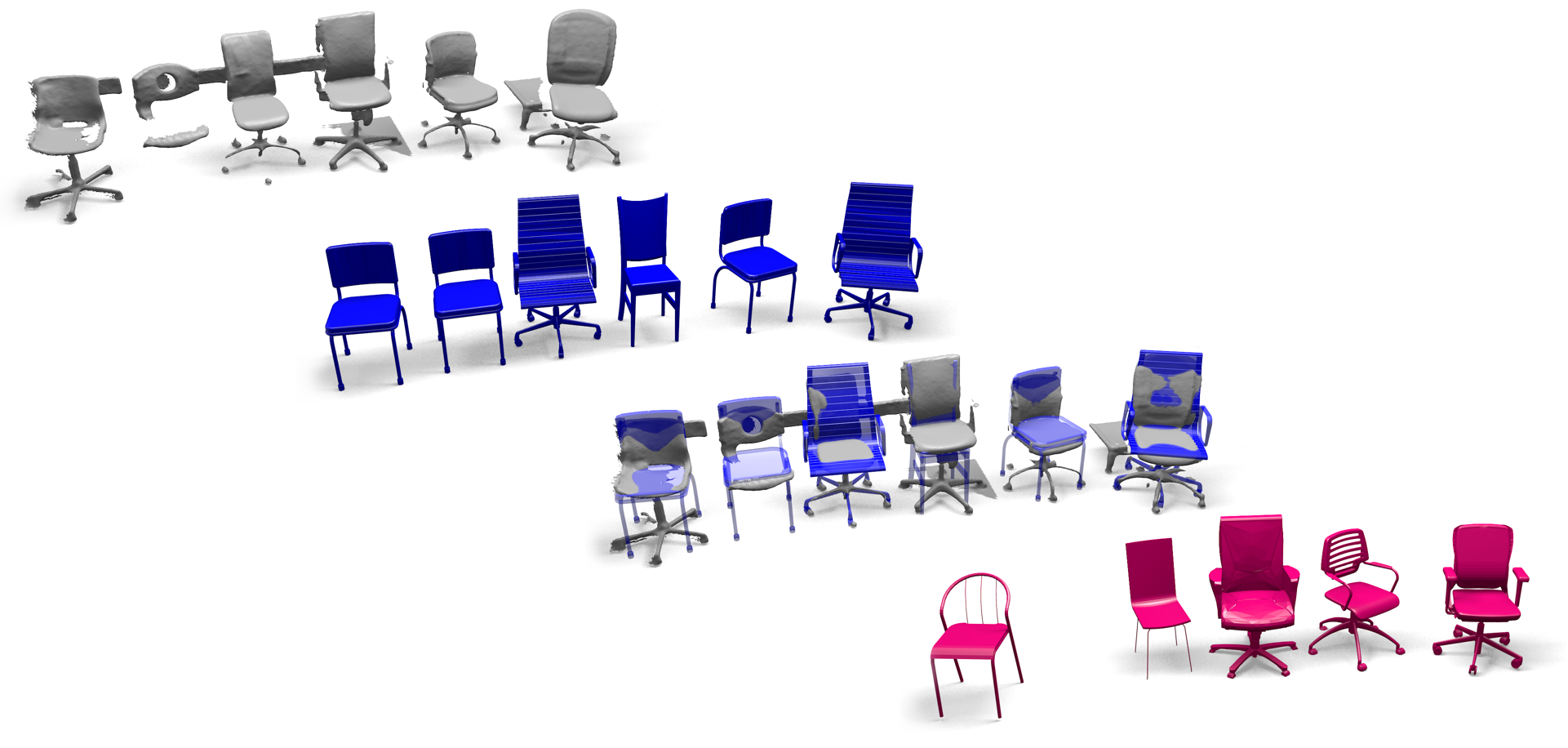}} \\

			\includegraphics[width = 0.23\textwidth]{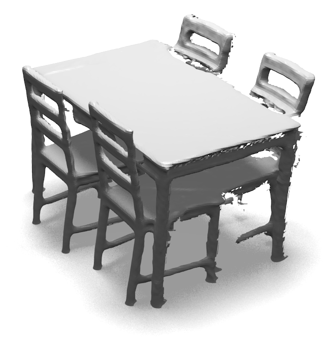} &		
			\includegraphics[width = 0.23\textwidth]{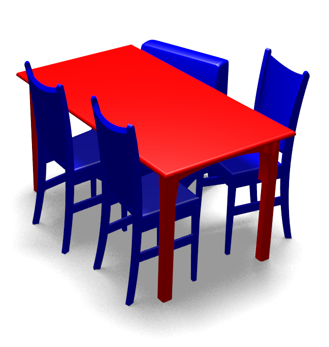} &
			\includegraphics[width = 0.23\textwidth]{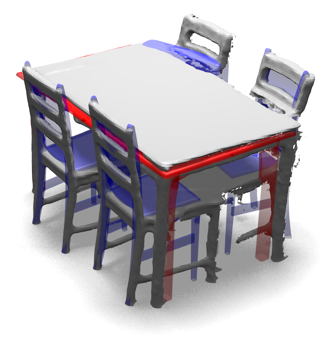} &
			\includegraphics[width = 0.23\textwidth]{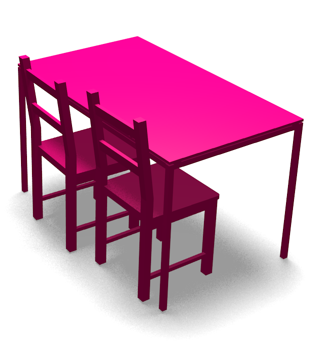} \\	
			
			(a) & (b) & (c) & (d) \\
		\end{tabular}   \\
		\caption{\textbf{Scenes from Li \etal\protect\cite{Li2015}.} (a) Scanned scene; (b) ASIST result; (c)  result overlay; (d) Li \etal result, shown in pink. The images of the top example are presented diagonally while the images of the bottom example are presented in a horizontal row.}		
	\end{figure*}

	In this section we compare the performance of ASIST to the algorithm introduced in \cite{Li2015}. Unfortunately, the authors provide no basis for quantitative comparison, due to the lack of annotated scenes and code. However, some of the scans were released along with their reconstructions, and we were therefore able to provide a qualitative comparison. Figure \ref{ikea_scenes} shows a comparison between the performance of ASIST (column (b)) and the published reconstruction of Li \etal (column (d)) on two different scenes. For both scenes it can be seen that ASIST achieves a perfect result in terms of both semantic and geometric recall and precision and thus outperforms the results of Li \etal In particular, it can be seen in the top scene that the solution of Li \etal misses one of the six chairs, while in the bottom scene it misses two of the four chairs.  These errors occur at quite challenging parts of the scans: the scanned chairs are missing significant chunks due to occlusions. Nevertheless, ASIST still performs well, inserting chairs appropriately.
	
	Furthermore, it is apparent that the reconstruction stays loyal to the geometry and the pose of the objects in the scene (coloumn (c)). A close look at the results of both algorithms shows that in some of the chairs, for instance, the reconstructed object is a four-legged chair whereas the object in the scene is a swivel chair. In ASIST, this is mainly due to the lack of diversity of exemplars. The set of exemplars we picked contains a single exemplar representing a ``swivel chair'' with a high backrest, hence for objects such as the second chair from the right and the leftmost one, a lower energy value can be obtained by choosing a chair with higher geometric error in the legs area than in the backrest area.
	
	%\begin{enumerate}
	%	\item Images of several sections of our results where successful, on scenes from Li \etal. \cite{Li2015}. \orl{DONE}
	%	\item Images of the same sections from Li's own results. \orl{DONE}
	%	\item (IF POSSIBLE) Quantitative comparison with Li: Table with precision/recall (geometric and semantic) of both our algorithm and Li's results. \orl{I think there are not enough results to do that}
	%\end{enumerate}
	
	\subsection{Comparison to Nan \etal\cite{Nan2012}}
	\label{Nan_Results}
\begin{figure*}
		\label{Nan_scenes}
		\centering
		\begin{tabular}{ c c c c}
			\includegraphics[width = 0.23\textwidth]{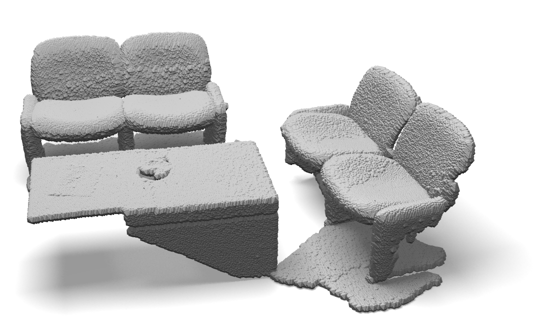} &
			\includegraphics[width = 0.23\textwidth]{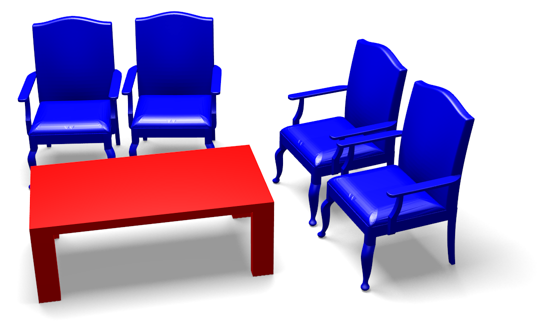} &
			\includegraphics[width = 0.23\textwidth]{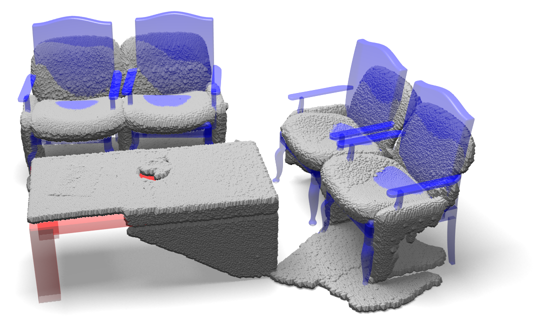} &
			\includegraphics[width = 0.23\textwidth]{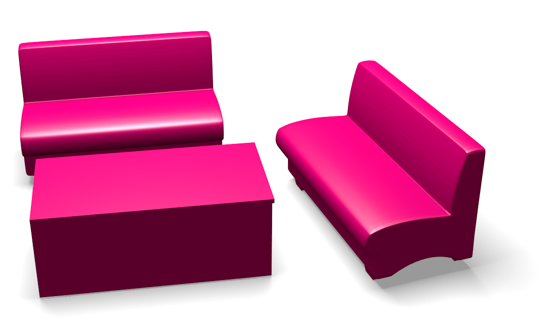} \\
			
			\includegraphics[width = 0.23\textwidth]{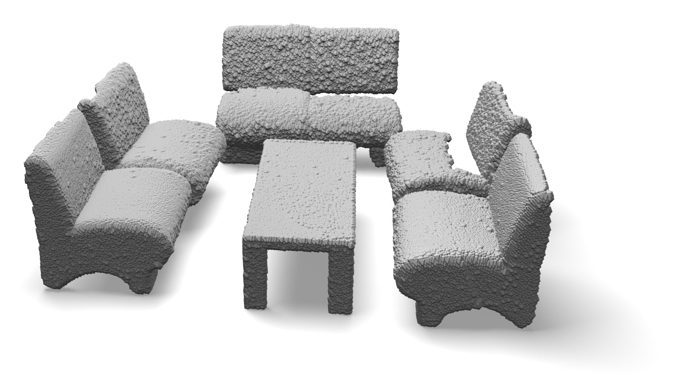} &
			\includegraphics[width = 0.23\textwidth]{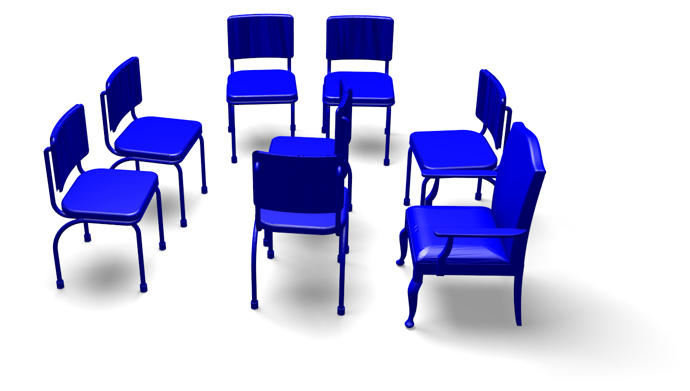} &
			\includegraphics[width = 0.23\textwidth]{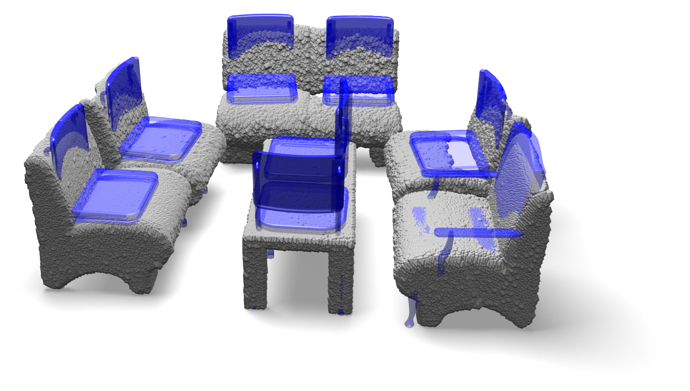} &
			\includegraphics[width = 0.23\textwidth]{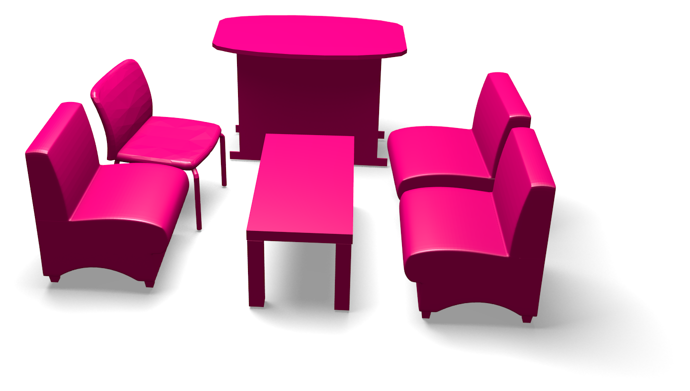} \\
			
			(a) & (b) & (c) & (d) \\
			
		\end{tabular}   \\
		\caption{\textbf{Scenes from Nan \etal\protect\cite{Nan2012}.} (a) Scanned scene; (b) ASIST result; (c)  result overlay; (d) Nan \etal result, shown in pink.}		
	\end{figure*}

	A quantitative comparison of ASIST to the algorithm proposed by Nan \etal \cite{Nan2012} is reported in Table \ref{table_Nan_PR}, and a qualitative comparison is shown in Figure \ref{Nan_scenes}. Since code was neither published nor was made available upon request, we evaluated the performance on the $18$ scene dataset that was released by the authors, and compared it against the score derived from their published reconstructions. In order to evaluate precision and recall we manually annotated bounding boxes of chairs, tables and sofas in the scenes. We intend to release this data to the research community. 
	
	Observe that while semantically, Nan \etal slightly outperforms ASIST in the majority of cases, the results are comparable both quantitatively and qualitatively (see Table \ref{table_Nan_PR}). Figure \ref{Nan_scenes} shows two of Nan's published scenes (column (a)), the result of ASIST (column (b)) and Nan's result (column (d)). In the top row one can observe a semantic failure case by Nan where the algorithm replaces the adjacent chairs with sofas. ASIST, on the other hand, finds the individual chairs correctly. The bottom row shows a case where Nan's method performs well except for replacing two adjacent chairs with a high table. In that case, ASIST finds all the chairs in the scene correctly, but erroneously replaces the coffee table with a couple of chairs as well.
	
	It is worthwhile noting that while Nan \etal allows deformations of the objects used for the reconstruction, ASIST uses its available exemplars as is. This results in a match which, while performing quite well in terms of precision and recall, deviates geometrically from the ground truth more than Nan's method. This provides the inspiration for one of the main directions for future research, as discussed in detail in Section \ref{sec:Discussion}.
	
	\begin{table}
		\begin{center}
			\begin{tabular}{| l | c | c | c |}
				\hline
				& chair       & table         & geo           \\
				\hline
				\textbf{Precision}     &             &               &      \\
				\hspace{1em} ASIST 		   & 0.91        & 0.8           & \tbf{0.93} \\
				\hspace{1em} Nan \etal      & \tbf{0.97}  & \tbf{0.81}    & 0.93       \\         
				\hline
				\textbf{Recall} & & & \\
				\hspace{1em} ASIST        & \tbf{0.93}  & 0.89          & \tbf{0.97}    \\
				\hspace{1em} Nan \etal    & 0.91        & \tbf{1}       & 0.93          \\
				\hline
				\textbf{$F_1$ score} & & & \\
				\hspace{1em} ASIST             & 0.92        & 0.84          & \tbf{0.95}    \\
				\hspace{1em} Nan \etal         & \tbf{0.94}  & \tbf{0.89}    & \tbf{0.93}     \\
				\hline
			\end{tabular}
			\caption{\textbf{Performance comparison with Nan \etal\protect\cite{Nan2012}}. Semantic and geometric precision, recall and $F_1$ scores for all scenes published in \protect\cite{Nan2012}.}
			\label{table_Nan_PR}
		\end{center}
	\end{table}

	\section{Discussion}
	\label{sec:Discussion}
	We have seen that the proposed ASIST algorithm is quite effective in achieving semantically invariant scene transformations on a variety of different sources of data.  However, in examining the failure cases, it becomes apparent that there are two elements of the algorithm which can lead ASIST astray.
	
	The first element is the cell classification.  We have observed that the random forest classifier has reasonable performance, but does not produce extraordinarily accurate results.  It turns out that this reasonable level of performance is sufficient for ASIST in many cases, as the other energy terms pull the algorithm towards the correct solution.  However, we observed that in many failure cases, it was the forest that provided a signal which was too weak, or even incorrect.  The semantic data term was consequently uninformative or incorrect, and the other energy terms could not adequately compensate.
	
	A potential solution to this problem is readily apparent.  Much work has been done on object recognition and detection; while the random forest algorithm we rely on is simple to implement, it is na\"{i}ve and has considerably lower performance than current state-of-the-art recognition algorithms, many of which are based on convolutional neural networks and other deep learning techniques.  For example, we might expect that using the detection algorithm introduced in Gupta \etal\cite{guptaECCV14}, the overall accuracy of the ASIST algorithm would improve.  One advantage of the way the ASIST pipeline is built is that plugging in such a state-of-the-art classifier is straightforward.
	
	The second element which negatively affects ASIST's performance relates to the registration step, i.e.,~the choice of the per exemplar transformations $T_e$.  In particular, in the current implementation we restrict ourselves to rigid transformations.  However, recall that our set of exemplars $\mathcal{E}$ is finite, and relatively small in practice.  Thus, it is often the case that an object in the scene will not perfectly match an exemplar, even if the best possible rigid transformation is chosen.  Consequently, it is sometimes the case that the resulting match is quite inaccurate.
	
	The issue is that the set of rigid transformations is too restrictive.  A remedy is to broaden the set of transformations to include scaling, possibly anisotropic.  Further afield, one might consider various classes of non-rigid transformations, for example of the type described in \cite{lombaert2014spectral}.  By expanding the set of transformations, one would expect more accurate matches with the scene. And while the computation due to a broader set of transformations might be more expensive, this could be offset by the need to use fewer exemplars in order to achieve accurate matching.
	
	\section{Conclusions}
	\label{sec:Conclusions}
	We have presented the ASIST algorithm for computing semantically invariant scene transformations.  Due to a unified formulation of semantic segmentation and object replacement based on the optimization of a single objective, ASIST solves both problems simultaneously via an iterative algorithm.  The method has been shown to achieve a high level of accuracy on datasets of both synthetic scenes and fused scans, as well as comparable performance to recently published competitor methods \cite{Nan2012,Li2015} on their own data.
	
	
	
	\section*{References}
	\bibliographystyle{elsarticle-num}
	\bibliography{ASIST_Bib}

\end{document}